\documentclass[10pt,twocolumn,letterpaper]{article}

\usepackage{cvpr}              

\usepackage{graphicx}
\usepackage{amsmath}
\usepackage{amssymb}
\usepackage{booktabs}
\usepackage{xcolor}
\usepackage{multirow}
\usepackage{placeins}
\usepackage{float}
\usepackage[pagebackref,breaklinks,colorlinks]{hyperref}

\usepackage[capitalize]{cleveref}
\crefname{section}{Sec.}{Secs.}
\Crefname{section}{Section}{Sections}
\Crefname{table}{Table}{Tables}
\crefname{table}{Tab.}{Tabs.}

\def\cvprPaperID{8867} 
\def\confName{CVPR}
\def\confYear{2023}

\begin{document}

\title{Re-IQA: Unsupervised Learning for Image Quality Assessment in the Wild }


\author{Avinab Saha$^*$ \quad \quad Sandeep Mishra$^*$ \quad \quad  Alan C. Bovik \\
Laboratory of Image and Video Engineering, The University of Texas at Austin 
}


\maketitle
\def\thefootnote{*}\footnotetext{Equal Contribution, Correspondence to Avinab Saha (avinab.saha@utexas.edu) \& Sandeep Mishra (sandy.mishra@utexas.edu). This work was supported by the National Science Foundation AI Institute for Foundations of Machine Learning (IFML) under Grant 2019844.  }

\begin{abstract}
  Automatic Perceptual Image Quality Assessment is a challenging problem that impacts billions of internet, and social media users daily. To advance research in this field, we propose a Mixture of Experts approach to train two separate encoders to learn high-level content and low-level image quality features in an unsupervised setting. The unique novelty of our approach is its ability to generate low-level representations of image quality that are complementary to high-level features representing image content. We refer to the framework used to train the two encoders as Re-IQA. For Image Quality Assessment in the Wild, we deploy the complementary low and high-level image representations obtained from the Re-IQA framework to train a linear regression model, which is used to map the image representations to the ground truth quality scores, refer Figure \ref{fig:architectures}. Our method achieves state-of-the-art performance on multiple large-scale image quality assessment databases containing both real and synthetic distortions, demonstrating how deep neural networks can be trained in an unsupervised setting to produce perceptually relevant representations. We conclude from our experiments that the low and high-level features obtained are indeed complementary and positively impact the performance of the linear regressor. A public release of all the codes associated with this work will be made available on GitHub. 
\end{abstract}

\section{Introduction}

\label{sec:intro}
 \begin{figure}[t!]
    \centering
    {{\includegraphics[width=1\linewidth]{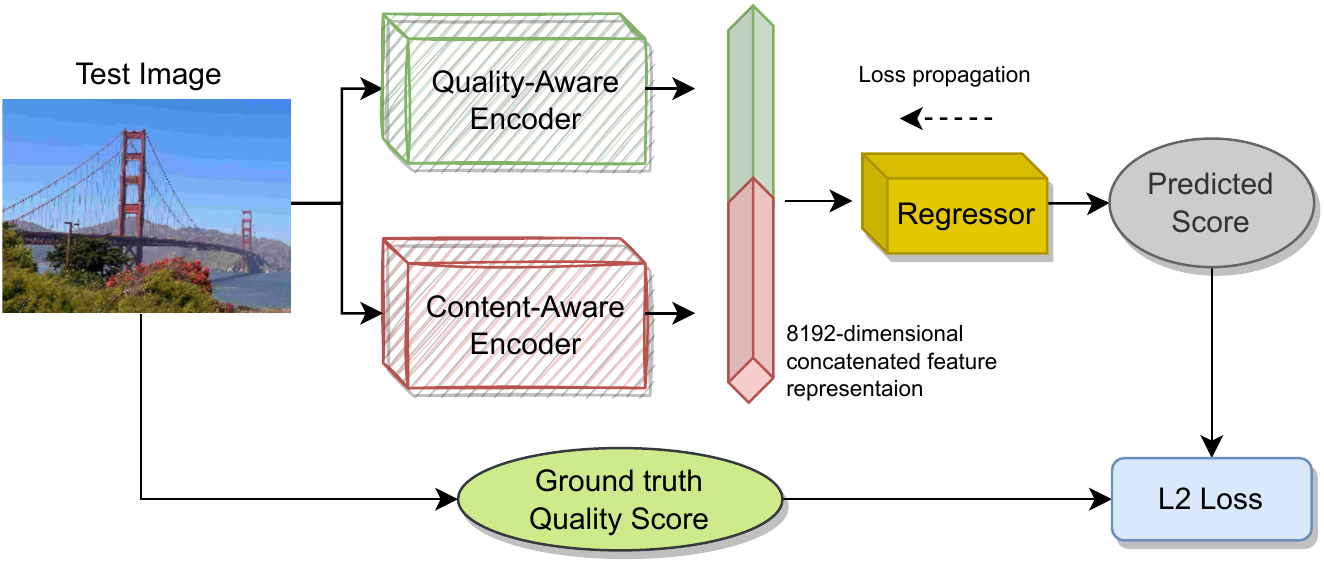}}}
    \captionsetup{justification=justified}
    \caption{IQA score prediction uses two encoders trained for complementary tasks of learning content and quality aware image representations. The encoders are frozen while the regressor learns to map image representations to quality predictions.}
    \label{fig:architectures}
\end{figure}

\begin{figure*}[htbp]
    \centering
    \subfloat[JPEG Compressed : 1 ]{{\includegraphics[width=0.24\textwidth]{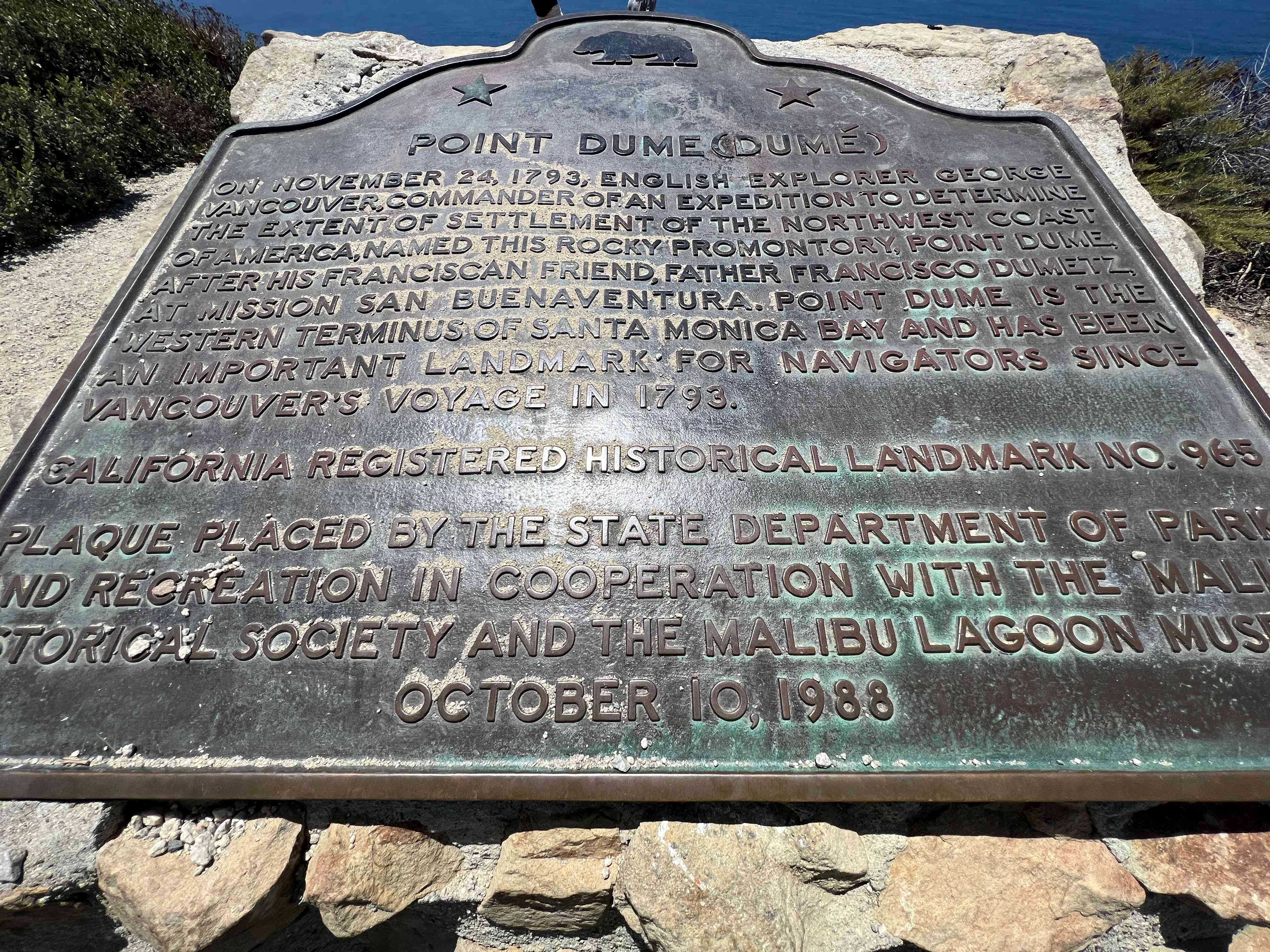}}}
    \hfill
    \subfloat[JPEG Compressed : 2 ]{{\includegraphics[width=0.24\textwidth]{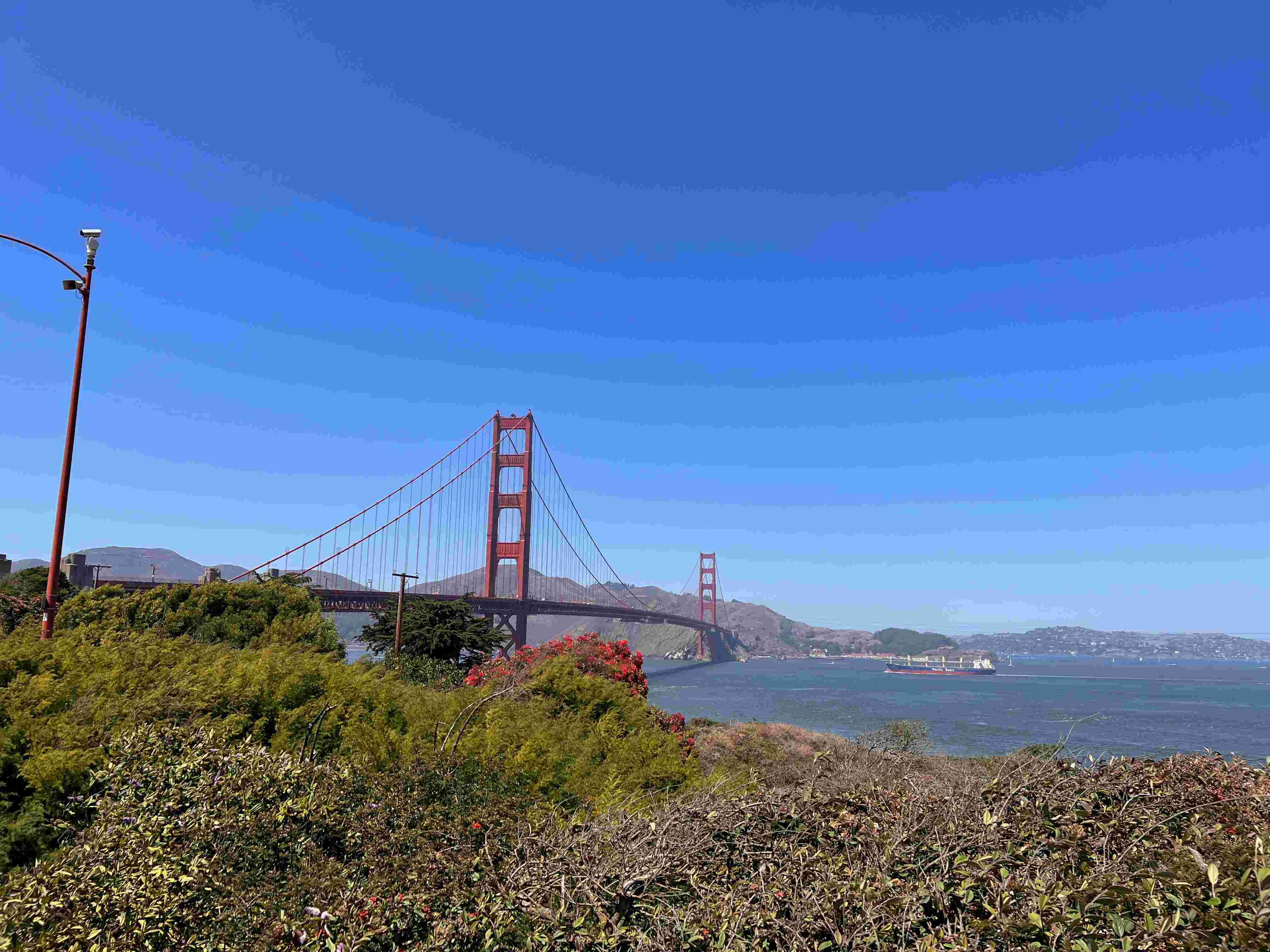}}}
    \hfill
     \subfloat[Motion Blur - Camera Shake ]{{\includegraphics[width=0.24\textwidth]{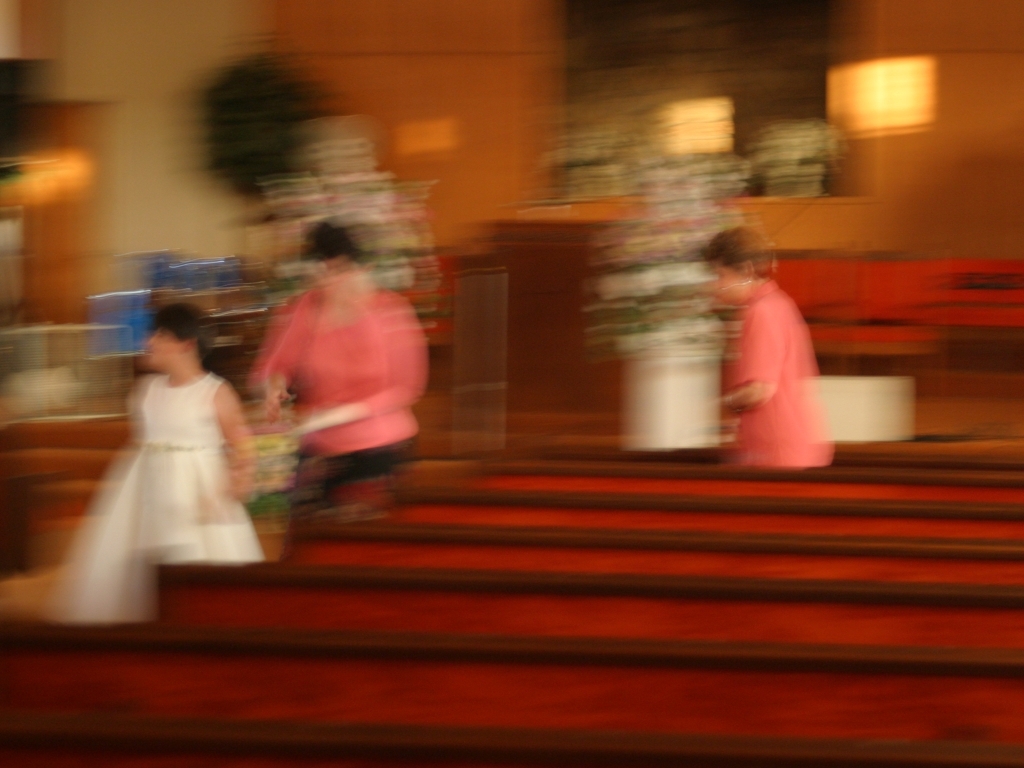}}}
    \hfill
    \subfloat[Overlaid Film Grain/ Noise]{{\includegraphics[width=0.24\textwidth]{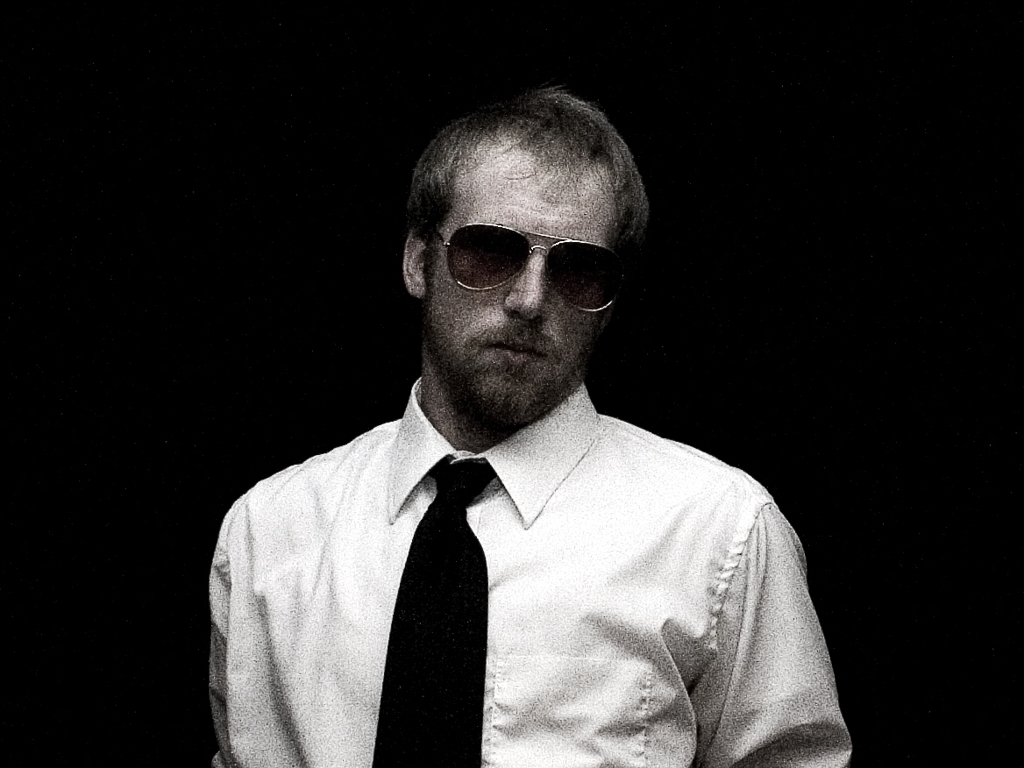}}}

    \captionsetup{justification=justified}
    \caption{Exemplar Synthetically and ``In the Wild" distorted pictures. (a), (b) are two images captured on iPhone 13 Pro and then JPEG compressed using the same encoding parameters. (c), (d), were taken from KonIQ and AVA datasets respectively, and exhibit typical ``Images in the Wild" distortions. Best viewed when zoomed in.}
    \label{fig:land}
\end{figure*}

Millions of digital images are shared daily on social media platforms such as Instagram, Snapchat, Flickr, etc. Making robust and accurate Image Quality Assessments (IQA) that correlate well with human perceptual judgments is essential to ensuring acceptable levels of visual experience. Social media platforms also use IQA metrics to decide parameter settings for post-upload processing of the images, such as resizing, compression, enhancement, etc. In addition, predictions generated by IQA algorithms are often used as input to recommendation engines on social media platforms to generate user feeds and responses to search queries. Thus, accurately predicting the perceptual quality of digital images is a high-stakes endeavor affecting the way billions of images are stored, processed, and displayed to the public at large. \\
\indent IQA metrics can be simply categorized into Full-Reference (FR) and No-Reference (NR) algorithms. FR-IQA algorithms like SSIM \cite{1284395}, FSIM \cite{5705575}, and LPIPS \cite{8578166} require both reference (undistorted) and distorted version of an image to quantify the human-perceivable quality. This requirement limits their applicability for the ``Images in the Wild" scenario, where the reference image is unavailable. On the contrary, NR-IQA algorithms like BRISQUE \cite{mittal2012no}, PaQ-2-PiQ \cite{DBLP:journals/corr/abs-1912-10088}, and CONTRIQUE \cite{DBLP:journals/corr/abs-2110-13266} do not require a reference image nor any knowledge about the kind of present distortions to quantify human-perceivable quality in a test image, paving the way for their use in ``Images in the Wild" scenarios.  \\ 
\indent No-Reference IQA for ``Images in the Wild" presents exciting challenges due to the complex interplay among the various kinds of distortions. Furthermore, due to the intricate nature of the human visual system, image content affects quality perception. In this work, we aim to learn low-level quality-aware image representations that are complementary to high-level features representative of image content. Figure \ref{fig:land} illustrates some of the challenges encountered in the development of NR-IQA algorithms. Figures \ref{fig:land} (a-b) show two images captured by the authors on an iPhone 13 Pro and compressed using the same encoding parameters. While any distortions are almost negligible in Figure \ref{fig:land} (a), there are artifacts that are clearly visible in Figure \ref{fig:land} (b). As in these examples, it is well known that picture distortion perception is content dependent, and is heavily affected by content related perceptual processes like masking \cite{Bovik2013AutomaticPO}. Figures \ref{fig:land} (c-d) illustrates a few distorted pictures ``In the Wild". Figures \ref{fig:land} (c-d) show two exemplar distorted pictures, one impaired by motion blur (Figure \ref{fig:land} (c))  and the other by film grain noise (Figure \ref{fig:land} (d)).  It is also well established that perceived quality does not correlate well with image metadata like resolution, file size, color profile, or compression ratio \cite{4775883}. Because of all these factors and the essentially infinite diversity of picture distortions, accurate prediction of the perception of image quality remains a challenging task, despite its apparent simplicity, and hence research on this topic remains quite active \cite{1284395,1055250,1576816,mittal2012no,ye2012unsupervised,6873260,5705575,8578166,DBLP:journals/corr/abs-2110-13266}. \\
\indent Our work is inspired by the success of momentum contrastive learning methods \cite{9157636,DBLP:journals/corr/abs-2003-04297} in learning unsupervised representations for image classification. In this work, we engineer our models to learn content and quality-aware image representations for NR-IQA on real, authentically distorted pictures in an unsupervised setting. We adopt a Mixture of Experts approach to independently train two encoders, each of which accurately learns high-level content and low-level image quality features. We refer to the new framework as Re-IQA. The key contributions we make are as follows:
\begin{itemize}
\itemsep0em
\item We propose an unsupervised low-level image quality representation learning framework that generates features complementary to high-level representations of image content. We demonstrate how the ``Mixture" of the two enables Re-IQA to produce image quality predictions that are highly competitive with existing state-of-the-art traditional, CNN, and Transformer based NR-IQA models, developed in both supervised and unsupervised settings across several databases.  

\item We demonstrate the superiority of high-level representations of image content for the NR-IQA task, obtained from the unsupervised pre-training of the ResNet-50 \cite{he2016deep} encoder over the features obtained from supervised pre-trained ResNet-50 on the ImageNet database \cite{Deng2009ImageNetAL}. We learn these high-level representations of image content using the unsupervised training framework proposed in MoCo-v2 \cite{DBLP:journals/corr/abs-2003-04297}
\item Inspired by the principles of visual distortion perception we propose a novel Image Augmentation and Intra-Pair Image Swapping scheme to enable learning of low-level image quality representations. The dynamic nature of the image augmentation scheme prevents the learning of discrete distortion classes, since it is applied to both pristine and authentically distorted images, enforcing learning of perceptually relevant image-quality features.

\end{itemize}

\section{Related Work}

\begin{figure*}[htbp]
    \centering
    \subfloat[Original Image ]{{\includegraphics[width=0.16\textwidth]{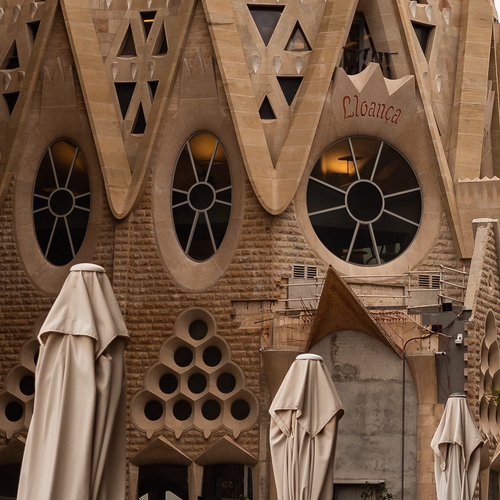}}}
    \hfill
    \subfloat[ Gaussian Blur ]{{\includegraphics[width=0.16\textwidth]{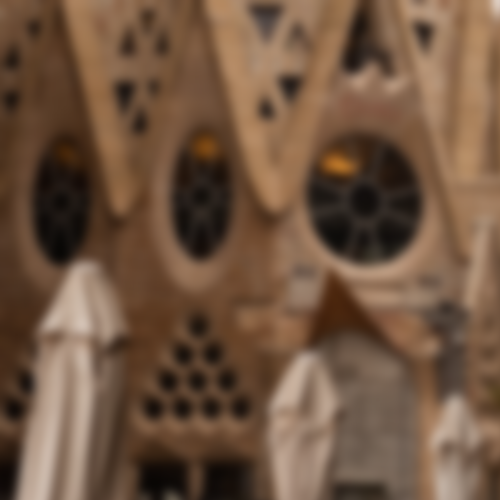}}}
    \hfill
    \subfloat[ Saturate ]{{\includegraphics[width=0.16\textwidth]{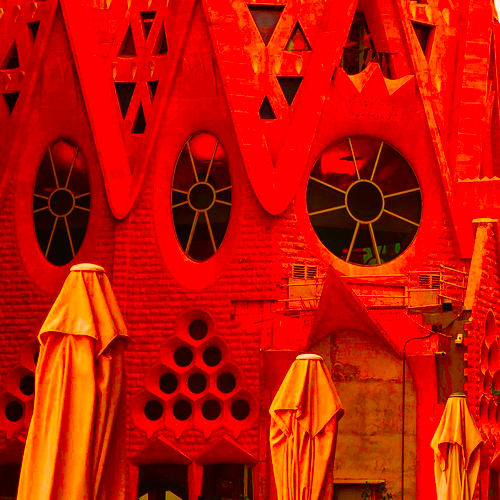}}}
    \hfill
    \subfloat[ Compression]{{\includegraphics[width=0.16\textwidth]{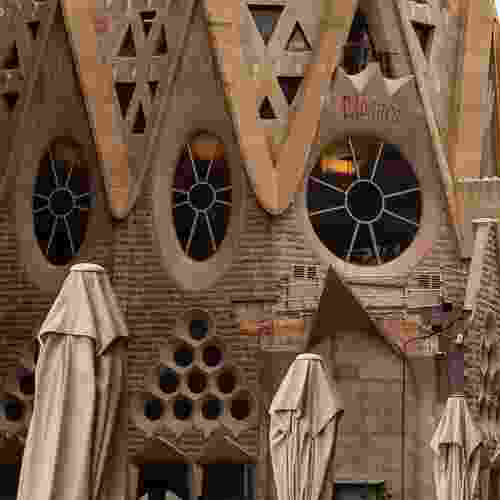}}}
    \hfill
    \subfloat[ Gaussian Noise]{{\includegraphics[width=0.16\textwidth]{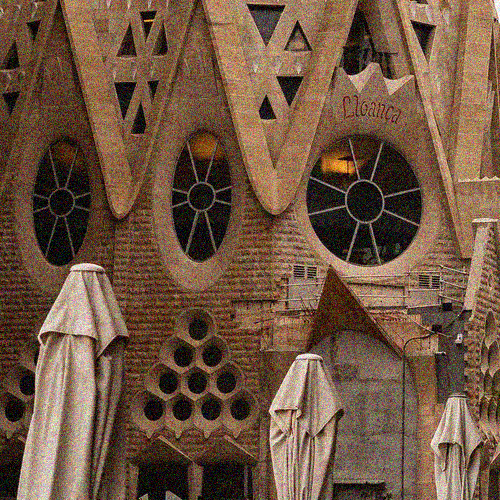}}}
    \hfill
    \subfloat[ Color Saturate]{{\includegraphics[width=0.16\textwidth]{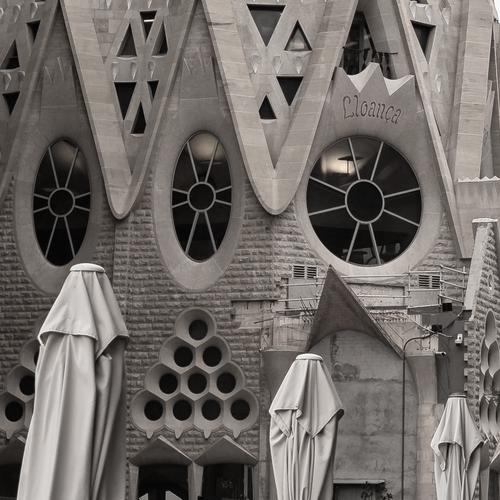}}}

    \subfloat[ Pixelate ]{{\includegraphics[width=0.16\textwidth]{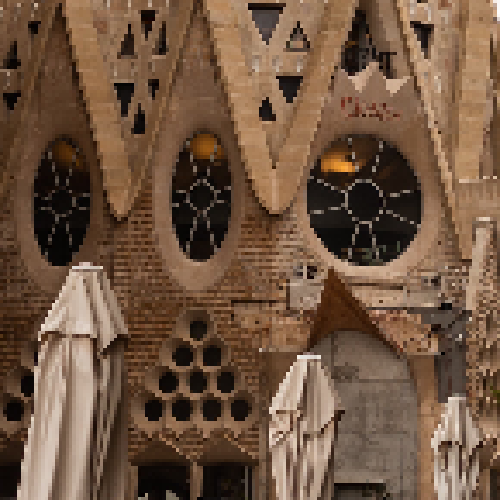}}}
    \hfill
    \subfloat[ Contrast]{{\includegraphics[width=0.16\textwidth]{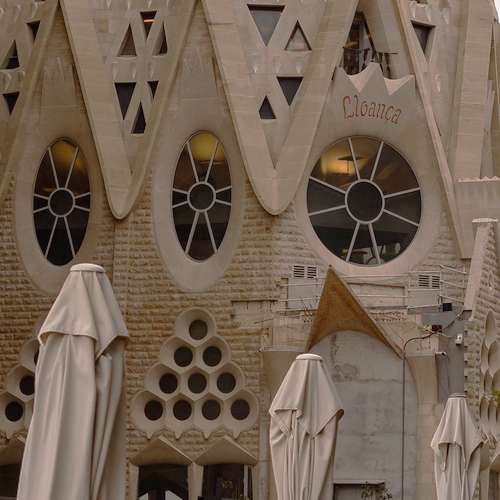}}}
    \hfill
    \subfloat[ Darken]{{\includegraphics[width=0.16\textwidth]{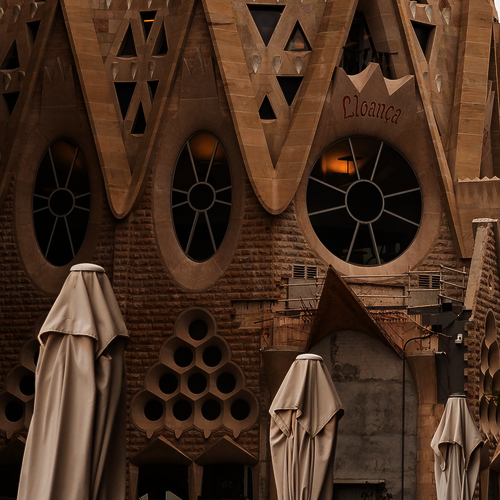}}}
    \hfill
    \subfloat[ Resizing]{{\includegraphics[width=0.16\textwidth]{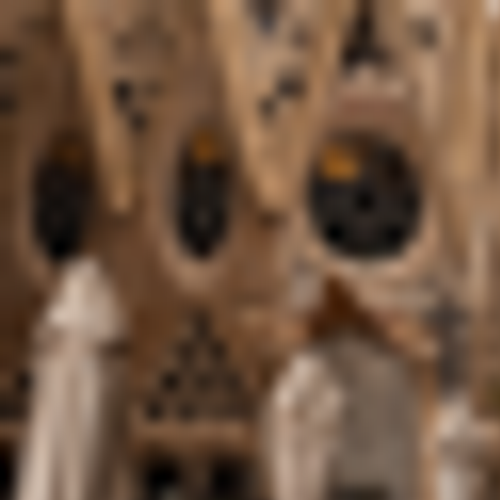}}} 
    \hfill
    \subfloat[ Jitter]{{\includegraphics[width=0.16\textwidth]{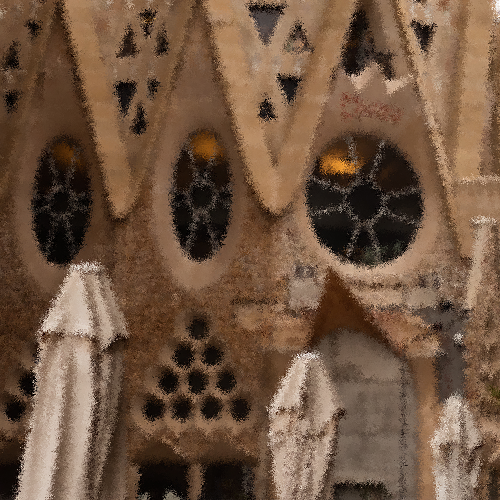}}} 
    \hfill
    \subfloat[ Motion Blur]{{\includegraphics[width=0.16\textwidth]{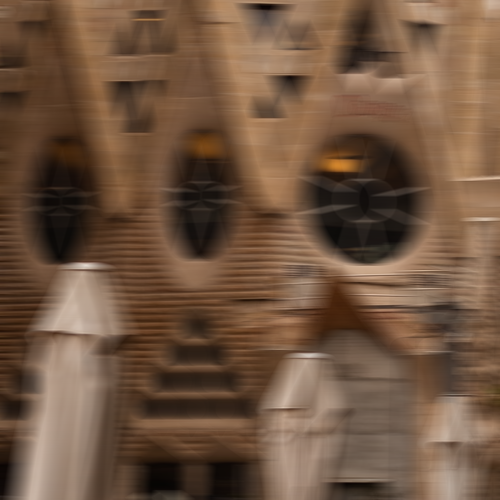}}} 
    
    \captionsetup{justification=justified}
    \caption{Some samples of distortions available in the Image Augmentation Scheme. There are a total of 25 distortions available in the bank, with 5 levels of distortion for each. More details are provided in Section \ref{ssec:dbanks}.}
    \label{fig:land1}
\end{figure*}
As discussed in Section \ref{sec:intro}, perceptual image quality prediction for ``Images in the Wild" is a challenging task due to the complex distortions that arise, and the combinations of them, and how they are perceived when they affect different kinds of pictorial content. Over the last few decades, a great deal of effort has been invested in the development of NR-IQA models that are able to accurately predict human judgment of picture quality. In recent years, NR-IQA models have evolved from using hand-crafted perceptual features, feeding shallow learners, into Deep Learning based approaches trained on large subjective databases. Traditional NR-IQA models generally have two components: a feature extractor, which generates quality-relevant features, and a low-complexity regression model, which maps the extracted features to quality scores. Most prior models have focused on improving the feature extractor and, thus improving the performance of the overall IQA algorithm. A common practice in traditional NR-IQA methods is to model image artifacts using statistical information extracted from a test image. Natural Scene Statistics (NSS) models and distorted versions of them are popular, where features are extracted from transformed domains, on which statistical measurements of deviations due to distortions are used as features for NR-IQA. For example, the NSS-based BRISQUE\cite{mittal2012no} and NIQE\cite{mittal2012making} models obtain features that capture in a normalized bandpass space \cite{mscn}. DIIVINE\cite{moorthy2011blind} uses steerable pyramids, and BLIINDS\cite{saad2012blind} uses DCT coefficients to measure statistical traces of distortions. Other methods like CORNIA\cite{ye2012unsupervised} and HOSA\cite{xu2016blind} utilize codebooks constructed from local patches, which are applied to obtain quality-aware features. Most of the methods discussed above often obtain acceptable results when evaluated on synthetically distorted images, but their performances significantly degrade when applied to ``Images in the Wild". This is because the above-discussed methods focus primarily on modeling the distortions present in a test image as statistical deviations from naturalness while completely ignoring the high-level content present in the image. \\
\indent The majority of deep learning approaches utilize pre-trained CNN backbones as feature extractors. This is done since end-to-end supervised training of NR-IQA models is difficult given the limited sizes of existing perceptual quality databases. These models typically use CNN backbones trained for ImageNet classification to extract features, combining them with low-complexity regression models like Support Vector Regression or Linear Regression to map the features to human-labeled scores. A few models use labeled scores from IQA databases to fine-tune the CNN backbone. The authors of the RAPIQUE \cite{tu2021rapique} show that features obtained from pretrained ResNet-50 \cite{he2016deep} could effectively predict quality scores on  ``In the Wild" content. In DB-CNN \cite{zhang2018blind}, authors adopt a two-path technique, where one CNN branch generates distortion-class and distortion-level features, while the other CNN branch provides high-level image content information. These are then combined using bilinear pooling. PQR \cite{zeng2017probabilistic} achieved faster convergence and better quality estimates by using the statistical distributions of subjective opinion scores instead of just scalar mean opinion scores (MOS) during training. BIECON \cite{kim2016fully} trains a CNN model on patches of distorted images, using proxy quality scores generated by FR-IQA models as labels. The authors of \cite{9156687} proposed an adaptive hyper-network architecture that considers content comprehension during perceptual quality prediction. Very recent works on NR-IQA includes PaQ-2-PiQ \cite{DBLP:journals/corr/abs-1912-10088}, CONTRIQUE \cite{DBLP:journals/corr/abs-2110-13266} and MUSIQ  \cite{DBLP:journals/corr/abs-2108-05997}. PaQ-2-PiQ benefits from a specially designed dataset wherein the authors not only collected subjective quality scores on whole pictures but also on a large number of image patches. The dataset is also large enough to train deep models in a supervised setting,  and PaQ-2-PiQ achieves state-of-the-art performance. However, although the authors use patch-level and image-level quality information during training, the training process may be susceptible to dataset sampling errors since only a few patches were extracted from each image and annotated with quality scores. MUSIQ uses a transformer-based architecture \cite{vaswani} pre-trained on the ImageNet classification dataset. The method benefits significantly by using transformer architecture and fine-tuning the transformer backbone on the IQA test databases. CONTRIQUE is a closely related work that aims to learn quality-aware representations in a self-supervised setting. CONTRIQUE learns to group images with similar types and distortion degrees into classes on an independent dataset. In this way, it proposes to learn quality-aware image representations. However, the class labels used to define `similar-quality' and `different-quality' samples in CONTRIQUE are in fact distortion labels instead of being true-quality labels. We address this shortcoming in our proposed framework. Our method, which is also completely unsupervised, does not learn representations based on distortion class labels, which can be inaccurate when asserted on ``In the Wild" data. Instead, our model, which is inspired by the fundamental principles of visual distortion perception, proposes to independently learn high-level semantic image content and low-level image quality features. These image representation features are mapped directly to subjective scores using a low-complexity regression model without fine-tuning the deep neural networks. Here, we utilize ResNet-based architectures \cite{he2016deep} throughout the Re-IQA framework. As our proposed framework is generalizable enough to be implemented using other CNN and transformer-based architectures, we plan to extend it to transformer-based architectures in the future.

\section{Rethinking-IQA}
\label{sec:method}
\begin{figure*}[htbp]
    \centering
    {{\includegraphics[width=1\linewidth]{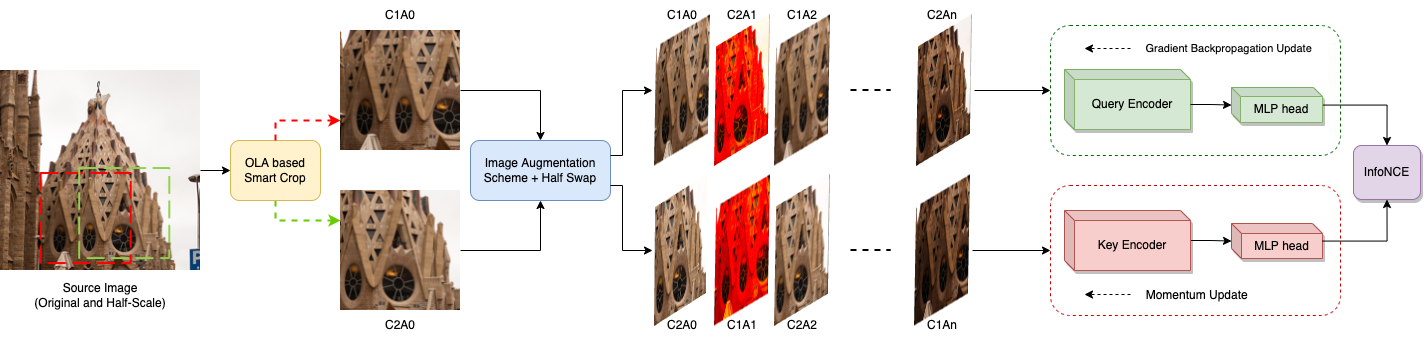}}}
    \captionsetup{justification=justified}
    \caption{Learning Quality Aware Representations: The OLA based cropping, Image Augmentation scheme and Half-swapping enable the generation of appropriate `similar-quality' and `different-quality' image pairs which can be used to learn quality-aware features. Note that $A_0$ has no augmentation, while $A_1..A_n$ are randomly sampled from the augmentation bank. During loss calculation, representations generated using the key encoder for the previous 65536 samples are also used as negative keys, following MoCo-V2 settings. }
    \label{fig:REIQA_pipeline}
\end{figure*}

The Re-IQA model framework is embodied in three processing phases. The first and second phases consist of training two ResNet50 encoders using contrastive learning of high-level and low-level image information. We then use the pre-trained encoders with frozen weights as image representation generation backbones, which supply features to a low-complexity regression model that is trained to conduct image quality prediction as shown in Figure \ref{fig:architectures}.

\indent To learn high-level content-aware information we deploy MoCo-v2\cite{DBLP:journals/corr/abs-2003-04297} ImageNet pre-trained model and adopt the design considerations developed in the original paper. Further discussed in section \ref{ssec:CAIR}. To learn quality-aware representations we develop a contrastive learning framework that deploys  a novel augmentation protocol and an intra-pair image-swapping scheme to facilitate model convergence towards learning robust image quality-aware features. Further discussed in section \ref{ssec:QAIR}.

\subsection{Re-IQA : Content Aware} \label{ssec:CAIR}

The primary objective in the MoCo-v2 framework \cite{DBLP:journals/corr/abs-2003-04297} is to assign a `similar' label to two crops from a single image while assigning a `different' label to two crops taken from two different images. Although, content aware Re-IQA based completely on the original MoCo-v2 framework performs well in the image quality prediction problem (refer Table \ref{tab:resultsmain}), it still suffers from a critical design problem: two crops from a same image can be given significantly different quality scores by human viewers. Hence we only use, the original MoCo-v2 framework to generate content-aware image representations. We make appropriate changes, discussed next, to the MoCo-v2 framework to enable accurate learning of quality-aware representations that complement content-aware representations.

\subsection{Re-IQA : Quality Aware} \label{ssec:QAIR}
\indent Our quality-aware contrastive learning framework uses an Image Augmentation method and an Intra-pair image Swapping scheme to train a ResNet-50 encoder within the MoCo-v2 framework \cite{DBLP:journals/corr/abs-2003-04297} with a goal of modeling a feature space wherein all images having similar degrees of perceptual quality fall closer to one another than to images having different perceptual qualities. The MoCo-v2 framework simultaneously processes a query image through the query encoder and a key image through the key encoder. In a single batch, a positive sample occurs when features are generated using any paired query and key, the pair being labeled `similar.' A negative sample occurs when the query and the key do not belong to the same pair, hence they are marked `different.' \\
 \indent To train a contrastive network we need paired images, such that for any sample index $k$, we have image pairs $[{i{_1}}^k, {i{_2}}^k]$ that can be assigned the `similar-quality' label, and for any $j,k;$ where $ k \neq j$ we have image pairs $[{i{_1}}^k, {i{_2}}^j]$ that can be assigned the `different-quality' label. From here on we shall refer to perceptual quality-aware features as PQAF. To define the decision boundary between `similar-quality' and `different-quality' labels, we assume the following three hypotheses to be true:\\  
 \textbf{H1}: PQAF varies within an image itself. If we assign PQAF to an image patch $x$ and denote it as $PQAF_{x}$, then $PQAF_{x}$ varies only a small amount between neighboring patches. However, $PQAF_{x}$ may vary significantly between two distant patches.  \\
 \textbf{H2}: The PQAF of any two randomly selected images are `different,' which assumes that the scenes depicted in the images to be different. However, this does not enforce any restrictions on the quality scores of the two images. \\
\textbf{H3}: Two different distorted versions of the same image have different PQAF. \\
These hypotheses are further discussed in Section \ref{ssec:dischyp}.

\subsubsection{Quality-Aware Image Augmentation Scheme} \label{ssec:QAIAS}
 To conduct quality-aware image feature extraction, we deploy a novel bank of image quality distortion augmentations, as elaborated in Section \ref{ssec:dbanks}. The augmentation bank is a collection of 25 distortion methods, each realized at $5$ levels of severity. For any source image $i^k$ from the training set, where $k\in\{1,2...K\}$ and $K$ is the total number of images in the training data, a randomly chosen subset of the augmentations available in the bank are applied to each image resulting in a mini-batch of distorted images. We combine each source image with its distorted versions to form $chunk^k$:
 \begin{equation}
     chunk^k = [i^k,{i{_1}}^k, {i{_2}}^k,..., {i{_{n}}}^k]
 \end{equation}
where ${i{_j}}^k$ is the $j^{th}$ distorted version of $i^k$, and $n$ is the number of augmentations drawn from the bank. We then generate two random crops of $chunk^k$, namely $chunk^{k_{c1}}$ and $chunk^{k_{c2}}$, using an overlap area based smart cropping mechanism. We choose these crop locations such that the overlapping area (OLA) in the two crops falls within minimum and maximum bounds. We make sure that the crop location is the same over all images in each chunk and different between chunks, resulting in:

\begin{equation}
    \begin{split}
    chunk^{k_{c1}} &= [i^{k_{c1}},{i{_1}}^{k_{c1}}, {i{_2}}^{k_{c1}},..., {i{_{n}}}^{k_{c1}}] \\
    chunk^{k_{c2}} &= [{i}^{k_{c2}},{i{_1}}^{k_{c2}}, {i{_2}}^{k_{c2}},..., {i{_{n}}}^{k_{c2}}] 
    \end{split}\label{eq:label2}
\end{equation}
When training, by choosing an augmented image $a_{th}$ from both $chunk^{k_{c1}}$ and $chunk^{k_{c2}}$, form the pair $[i_a^{k_{c1}}, i_a^{k_{c2}}]$. Image $i_a^{k_{c1}}$ and $i_a^{k_{c2}}$ are neighboring patches because of OLA-based cropping and hence are marked `similar-quality' as stated in \textbf{H1}. Similarly, for any image $k$ and distortion $a, b$, where $a\neq b$, the pair $[i_{a}^{k_{c1}}, i_{b}^{k_{c2}}]$ are labelled as `different-quality' as in \textbf{H1}. Finally, 
 for any two different image samples $k,j$, label the pair $[i_{\_}^{k_{c1}}, i_{\_}^{j_{c2}}]$ as `different-quality', following \textbf{H2}. 
 
\subsubsection{Intra-Pair Image Swapping Scheme} \label{ssec:IPISS}

Given a spatial arrangement of $chunk^{k_{c1}}$ and $chunk^{k_{c2}}$:

\begin{gather*}
    \begin{split}
    &\textcolor{blue}{i^{k_{c1}}}\ \ \ \  \textcolor{blue}{{i{_1}}^{k_{c1}}}\ \ \ \   ...\ \ \ \ \textcolor{blue}{{i{_{m}}}^{k_{c1}}}\ \ \ \ \textcolor{blue}{{i{_{m+1}}}^{k_{c1}}}\ \ \ \ ...\ \ \ \ \textcolor{blue}{{i{_{n-1}}}^{k_{c1}}}\ \ \ \ \textcolor{blue}{{i{_{n}}}^{k_{c1}}} \\
    & \ \ \updownarrow \ \ \ \ \ \ \ \updownarrow \ \ \ \ \ \ \ \ \ \ \ \ \ \ \ \ \ \ \updownarrow \ \ \ \ \ \ \ \ \ \ \ \updownarrow \ \ \ \ \ \ \ \ \ \ \ \ \ \ \ \ \ \ \ \ \ \ \ \updownarrow \ \ \ \ \ \ \ \ \ \ \ \ \ \updownarrow
    \\
    &\textcolor{red}{{i}^{k_{c2}}}\ \ \ \ \textcolor{red}{{i{_1}}^{k_{c2}}}\ \ \ \ ...\ \ \ \ \textcolor{red}{{i{_{m}}}^{k_{c2}}}\ \ \ \ \textcolor{red}{{i{_{m+1}}}^{k_{c2}}}\ \ \ \ ...\ \ \ \ \textcolor{red}{{i{_{n-1}}}^{k_{c2}}}\ \ \ \ \textcolor{red}{{i{_{n}}}^{k_{c2}}}
    \end{split}
\end{gather*}
form the following types of image-pairs and corresponding labels:
\begin{gather*}
    \begin{split}
        [\textcolor{blue}{i_{m}^{k_{c1}}}, \textcolor{red}{i_{m}^{k_{c2}}}] &\mapsto similar-quality\\
        [\textcolor{blue}{i_{m}^{k_{c1}}}, \textcolor{red}{i_{l}^{k_{c2}}}] &\mapsto different-quality
    \end{split}
\end{gather*}
Then apply intra-pair image swapping on the generated chunks to obtain the following arrangement:

\begin{gather*}
    \begin{split}
    &\textcolor{blue}{i^{k_{c1}}}\ \ \ \ \textcolor{red}{{i{_1}}^{k_{c2}}}\ \ \ \ ...\ \ \ \ \textcolor{blue}{{i{_{m}}}^{k_{c1}}}\ \ \ \ \textcolor{red}{{i{_{m+1}}}^{k_{c2}}}\ \ \ \ ...\ \ \ \ \textcolor{blue}{{i{_{n-1}}}^{k_{c1}}}\ \ \ \ \textcolor{red}{{i{_{n}}}^{k_{c2}}} \\
    & \ \updownarrow \ \ \ \ \ \ \ \ \updownarrow \ \ \ \ \ \ \ \ \ \ \ \ \ \ \ \ \ \ \updownarrow \ \ \ \ \ \ \ \ \ \ \ \ \updownarrow \ \ \ \ \ \ \ \ \ \ \ \ \ \ \ \ \ \ \ \ \ \ \updownarrow \ \ \ \ \ \ \ \ \ \ \ \ \ \ \ \updownarrow
    \\
    &\textcolor{red}{{i}^{k_{c2}}}\ \ \ \ \textcolor{blue}{{i{_1}}^{k_{c1}}}\ \ \ \ ...\ \ \ \ \textcolor{red}{{i{_{m}}}^{k_{c2}}}\ \ \ \ \textcolor{blue}{{i{_{m+1}}}^{k_{c1}}}\ \ \ \ ...\ \ \ \ \textcolor{red}{{i{_{n-1}}}^{k_{c2}}}\ \ \ \ \textcolor{blue}{{i{_{n}}}^{k_{c1}}}
    \end{split}
\end{gather*}

By swapping images within each pair over half the pairs, (referred to as Half Swap), the network is introduced to samples having the following configuration: $[{i_{a}^{k_{c1}}}, {i_{b}^{k_{c1}}}]$ where $a,b; a\neq b$ are two different distortions. Note that the crops $[i_a^{k_{c1}}, i_b^{k_{c1}}]$ are exactly the same, except for the distortion applied, and thus contain the same essential visual content. Despite this, we mark such samples as `different-quality' as stated in \textbf{H3}, thus forcing the network to look beyond content-dependent features. With this, we finally end up with the following image pairs and labels:
\begin{gather*}
    \begin{split}
        [\textcolor{blue}{i_{m}^{k_{c1}}}, \textcolor{red}{i_{m}^{k_{c2}}}] &\mapsto similar-quality\\
        [\textcolor{blue}{i_{m}^{k_{c1}}}, \textcolor{red}{i_{l}^{k_{c2}}}] &\mapsto different-quality\\
        [\textcolor{blue}{i_{m}^{k_{c1}}}, \textcolor{red}{i_{l}^{k_{c1}}}] &\mapsto different-quality\\
        [\textcolor{blue}{i_{m}^{k_{c1}}}, \textcolor{red}{i_{l}^{j_{c2}}}] &\mapsto different-quality
    \end{split}
\end{gather*}

\subsubsection{Quality-Aware Training} 
\label{ssec:QAT}

Define two identical encoders 1) Online Encoder (query encoder) and 2) Momentum Encoder (key encoder). Both encoders have ResNet-50 backbones and an MLP head to generate the final output embeddings from the ResNet features. Split the pairs designed in the previous step, passing the first image from each pair through the query encoder, and the other through the key encoder. To calculate the loss between the representation generated by the query and key encoder, we use the InfoNCE \cite{oord2018representation} loss function:
\begin{equation}
    \!\!\mathcal{L}_{q, k^+, \{k^-\}} = -\log \frac{\exp(q.k^+/\tau)}{\exp{(q.k^+/\tau)}+\sum\limits_{k^-}^{} \exp(q.k^-/\tau)}
\end{equation}

Here $q$ is the query image, $k^+$ is a positive sample (similar-quality), ${k^-}$ represnt negative samples (different-quality), and $\tau$ is a temperature hyper-parameter. This loss is then used to update the weights of the online encoder by back-propagation. The weights of the momentum encoder are updated using the weighted sum of its previous weights and the new weights of the online encoder. Formally denoting the parameters of the query encoder by $\theta_{q}$ and the parameters of the key encoder as $\theta_{k}$, update $\theta_{k}$ as:

\begin{equation}
    \theta_{k} \leftarrow m\theta_{k} + (1-m)\theta_{q}
\end{equation}

Here $m\in[0,1)$, is the momentum coefficient. Once the encoder pre-training has saturated the frozen ResNet-50 can be used as an encoder backbone for any downstream task associated with perceptual image quality.
\subsection{IQA Regression}
We concatenate the image representations obtained from the content and quality-aware encoders in the previous steps to train a regressor head to map the obtained features to the final perceptual image quality scores as shown in Figure \ref{fig:architectures}. In our experiments, we use a single-layer perceptron as the regressor head. It is important to note that we train only the low-complexity regressor head while evaluating our Re-IQA framework across multiple databases. Our method does not require us to fine-tune the feature extraction backbone(s) separately for each evaluation database as required in MUSIQ. 
\section{Experimental Results}

\subsection{Training Datasets} 
In the Re-IQA framework, two ResNet-50 encoders are trained to obtain high-level image content features and low-level image quality features. The encoder that  learns high-level image content features was trained on a subset of the ImageNet database \cite{Deng2009ImageNetAL} containing approximately 1.28 million images across 1000 classes. When training the encoder in an unsupervised setting, we discard the class label information and only use images without labels during the training process. \\
\indent To learn the low-level image quality features, we use a combination of pristine images and authentically distorted images as training data. The augmentation scheme (applied to all images in the dataset) ensures that the network learns how to differentiate between distortions when the semantic content in the image is the same. The presence of authentically distorted images in the dataset helps tune the model to accurately predict the quality of ``In the Wild" pictures.
\begin{itemize}
\itemsep0em
\item Pristine Images: We used the 140K pristine images in the KADIS dataset \cite{lin2019kadid}. We do not use the 700K distorted images available in the same dataset. The authors of KADIS did not provide subjective quality scores for any image in the dataset.
\item Authentically Distorted Images: We used the same combination of datasets as proposed in CONTRIQUE \cite{DBLP:journals/corr/abs-2110-13266} to form our distorted image set: (a) AVA \cite{Murray2012AVAAL} - 255K images, (b) COCO \cite{lin2014microsoft} - 330K images, (c) CERTH-Blur \cite{mavridaki2014no} - 2450 images, d) VOC \cite{everingham2010pascal} - 33K images
\end{itemize}

\subsection{Evaluation Datasets}
\label{ssec:eval}
Many previous IQA methods used legacy databases like LIVE IQA \cite{sheikh2006statistical}, TID-2008 \cite{ponomarenko2009tid2008}, TID-2013 \cite{ponomarenko2013color}, CSIQ-IQA \cite{larson2010most}, and KADID \cite{lin2019kadid} for development and evaluation purposes. However, these datasets contain only a small number ($\sim 25-100$) of pristine images synthetically distorted by various levels and types of single distortions. Hence, these datasets lack diversity and realism of content and distortion. Recently many ``In the Wild" datasets like KonIQ \cite{hosu2020koniq}, CLIVE \cite{ghadiyaram2015massive}, FLIVE \cite{DBLP:journals/corr/abs-1912-10088}, and SPAQ \cite{fang2020perceptual} have been developed and used by visual quality researchers since they address the shortcomings of the legacy datasets. The newer breed of perceptual quality datasets contain many authentically distorted images.  KonIQ-10K dataset consists of 10K images selected from the publicly available multimedia large-scale YFCC100M database \cite{thomee2016yfcc100m}. CLIVE contains 1162 authentically distorted images captured using various mobile devices. FLIVE, on the other hand, is comprised of 40,000 images sampled from open-source images and designed to emulate the feature distributions found in social media images statistically. Lastly, SPAQ consists of 11,000 images captured using 66 mobile devices, and each image is accompanied by various annotations such as brightness, content labels, and EXIF data. However, our experiments only utilize the image and its corresponding quality score.

\begin{table*}[]
\centering
\resizebox{\textwidth}{!}{%
\begin{tabular}{@{}c|c|ccccc|ccccc|cccc@{}}
\hline
\multirow{2}{*}{\begin{tabular}[c]{@{}c@{}}Model\\ \end{tabular}} &
  \multirow{2}{*}{\begin{tabular}[c]{@{}c@{}}Evaluation Dataset\\ \end{tabular}} &
  \multicolumn{5}{c|}{n\_aug} &
  \multicolumn{5}{c|}{Patch Size} &
  \multicolumn{4}{c}{OLA bound (\%)} \\ \cline{3-16}
 &      & 2 & 5 & 11 & 15 & 23 & 128 & 160 & 192 & 224 & 256 & 5-15 & 10-30 & 50-80 & No Bound \\ \cline{1-16} 
\multirow{3}{*}{\begin{tabular}[c]{@{}c@{}}Re-IQA\\ (Quality-Aware)\end{tabular}} &
  KonIQ & 0.855 & 0.858 & 0.861 & 0.862 & 0.860 & 0.860 & 0.861 & 0.861 & 0.860 & 0.858 & 0.856 & 0.861 & 0.857 & 0.858
   \\
 & SPAQ  & 0.890 & 0.896 & 0.900 & 0.900 & 0.897 & 0.898 & 0.900 & 0.901 & 0.898 & 0.894 & 0.889 & 0.900 & 0.886 & 0.890      \\
 & CSIQ  & 0.928 & 0.937 & 0.944 & 0.936 & 0.930 & 0.937 & 0.944 & 0.939 & 0.930 & 0.928 & 0.937 & 0.944 & 0.931 & 0.938      \\ \hline
\multirow{3}{*}{\begin{tabular}[c]{@{}c@{}}Re-IQA\\ (Quality-Aware +\\ Content Aware)\end{tabular}} &
  KonIQ & 0.895 & 0.901 & 0.914 & 0.898 & 0.895 & 0.910  & 0.914 & 0.911 & 0.905 & 0.897 & 0.903 & 0.914 & 0.897 & 0.905   
   \\
 & SPAQ  & 0.902 & 0.913 & 0.918 & 0.910  & 0.909 & 0.916 & 0.918 & 0.914 & 0.908 & 0.903 & 0.907 & 0.918 & 0.904 & 0.906   \\
 & CSIQ  & 0.932 & 0.941 & 0.947 & 0.937 & 0.935 & 0.940  & 0.947 & 0.942 & 0.937 & 0.932 & 0.939 & 0.947 & 0.935 & 0.940    \\ \bottomrule
\end{tabular}%
}
\caption{SRCC performance comparison of Re-IQA-QA and Re-IQA while varying one hyper-parameter at a time. While varying $n_{aug}$, we keep patch size $160$ and OLA bound $10-30\%$. When varying patch size, $n_{aug}$ was fixed to 11 and OLA bound to $10-30\%$. When varying OLA bound, $n_{aug}$ was set to 11 and the patch size was set to 160.}
\label{tab:ablationcombined}
\end{table*}

\begin{table*}[t]
\resizebox{\textwidth}{!}{%
\begin{tabular}{c|cccccccc|cccccccc}
\hline
\multirow{3}{*}{Method} &
  \multicolumn{8}{c|}{Authentic Distortions a.k.a ``Images in the Wild"} &
  \multicolumn{8}{c}{Synthetic Distortions} \\ \cline{2-17} 
 &
  \multicolumn{2}{c}{KonIQ} &
  
  \multicolumn{2}{c|}{CLIVE} &
  \multicolumn{2}{c}{FLIVE} &
  \multicolumn{2}{c|}{SPAQ} &
  \multicolumn{2}{c}{LIVE-IQA} &
  \multicolumn{2}{c|}{CSIQ-IQA} &
  \multicolumn{2}{c}{TID-2013} &
  \multicolumn{2}{c}{KADID} \\ \cline{2-17} 
 &
  SRCC &
  PLCC &
  SRCC &
  \multicolumn{1}{c|}{PLCC} &
  SRCC &
  PLCC &
  SRCC &
  PLCC &
  SRCC &
  PLCC &
  SRCC &
  \multicolumn{1}{c|}{PLCC} &
  SRCC &
  PLCC &
  SRCC &
  PLCC \\ \hline
BRISQUE &
  0.665 &
  0.681 &
  0.608 &
  \multicolumn{1}{c|}{0.629} &
  0.288 &
  0.373 &
  0.809 &
  0.817 &
  0.939 &
  0.935 &
  0.746 &
  \multicolumn{1}{c|}{0.829} &
  0.604 &
  0.694 &
  0.528 &
  0.567 \\
CORNIA &
  0.780 &
  0.795 &
  0.629 &
  \multicolumn{1}{c|}{0.671} &
  - &
  - &
  0.709 &
  0.725 &
  0.947 &
  0.950 &
  0.678 &
  \multicolumn{1}{c|}{0.776} &
  0.678 &
  0.768 &
  0.516 &
  0.558 \\
DB-CNN &
  0.875 &
  0.884 &
  0.851 &
  \multicolumn{1}{c|}{\textbf{0.869}} &
  0.554 &
  0.652 &
  0.911 &
  0.915 &
  0.968 &
  \textbf{0.971} &
  \textbf{0.946} &
  \multicolumn{1}{c|}{0.959} &
  0.816 &
  \textbf{0.865} &
  0.851 &
  0.856 \\
PQR &
  0.880 &
  0.884 &
  \textbf{0.857} &
  \multicolumn{1}{c|}{\textbf{0.882}} &
  - &
  - &
  - &
  - &
  0.965 &
  \textbf{0.971} &
  0.872 &
  \multicolumn{1}{c|}{0.901} &
  0.740 &
  0.798 &
  - &
  - \\
 
PaQ-2-PiQ &
  0.870 &
  0.880 &
  0.840 &
  \multicolumn{1}{c|}{0.850} &
  0.571 &
  0.623 &
  - &
  - &
  - &
  - &
  - &
  \multicolumn{1}{c|}{-} &
  - &
  - &
  - &
  - \\
HyperIQA &
  0.906 &
  0.917 &
  \textbf{0.859} &
  \multicolumn{1}{c|}{\textbf{0.882}} &
  0.535 &
  0.623 &
  0.916 &
  0.919 &
  0.962 &
  0.966 &
  0.923 &
  \multicolumn{1}{c|}{0.942} &
  0.840 &
  0.858 &
  0.852 &
  0.845 \\
CONTRIQUE &
  0.894 &
  0.906 &
  0.845 &
  \multicolumn{1}{c|}{0.857} &
  0.580 &
  0.641 &
  0.914 &
  0.919 &
  0.960 &
  0.961 &
  0.942 &
  \multicolumn{1}{c|}{0.955} &
  \textbf{0.843} &
  0.857 &
  \textbf{0.934} &
  \textbf{0.937} \\
  MUSIQ &
  \textbf{0.916} &
  \textbf{0.928} &
  - &
  \multicolumn{1}{c|}{-} &
  \textbf{0.646} &
  \textbf{0.739} &
  \textbf{0.917} &
  \textbf{0.921} &
  - &
  - &
  - &
  \multicolumn{1}{c|}{-} &
  - &
  - &
  - &
  - \\
ImageNet Pretrained (Supervised) &
  0.888 &
  0.904 &
  0.781 &
  \multicolumn{1}{c|}{0.809} &
  0.595 &
  0.648 &
  0.904 &
  0.909 &
  0.925 &
  0.931 &
  0.840 &
  \multicolumn{1}{c|}{0.848} &
  0.679 &
  0.729 &
  0.701 &
  0.677 \\ \hline
Re-IQA (content aware) &
  0.896 &
  0.912 &
  0.808 &
  \multicolumn{1}{c|}{0.844} &
  0.588 &
  0.699&
  0.902 &
  0.908 &
  0.867 &
  0.858 &
  0.766 &
  \multicolumn{1}{c|}{0.824} &
  0.658 &
  0.736 &
  0.601 &
  0.656 \\
Re-IQA (quality aware) &
  0.861 &
  0.885 &
  0.806 &
  \multicolumn{1}{c|}{0.824} &
  0.584 &
  0.590 &
  0.900 &
  0.910 &
  \textbf{0.971} &
  \textbf{0.972} &
  0.944 &
  \multicolumn{1}{c|}{\textbf{0.964}} &
  \textbf{0.844} &
  \textbf{0.880} &
  \textbf{0.885} &
  \textbf{0.892} \\
Re-IQA (content + quality) &
  \textbf{0.914} &
  \textbf{0.923} &
  0.840 &
  \multicolumn{1}{c|}{0.854} &
  \textbf{0.645} &
  \textbf{0.733} &
  \textbf{0.918} &
  \textbf{0.925} &
  \textbf{0.970} &
  \textbf{0.971} &
  \textbf{0.947} &
  \multicolumn{1}{c|}{\textbf{0.960}} &
  0.804 &
  0.861 &
  0.872 &
  0.885
\end{tabular}%
}
\caption{Performance comparison of Re-IQA against various NR-IQA models on IQA databases with authentic and synthetic distortions. The top 2 best performing models are in bold. Higher SRCC and PLCC imply better performance. MUSIQ results from \cite{DBLP:journals/corr/abs-2108-05997}. Results of all other existing methods from \cite{DBLP:journals/corr/abs-2110-13266}. }
\label{tab:resultsmain}
\end{table*}

We also evaluated our method on four legacy synthetically distorted datasets: LIVE-IQA, TID-2013, CSIQ-IQA, and KADID. We provide short descriptions of each of the databases. The LIVE IQA dataset includes 779 images that have been synthetically distorted using five types of distortions at four different intensity levels. TID-2013, on the other hand, contains 3000 images that have been synthetically distorted using 24 different types of distortions at five different levels of intensity, with 25 reference images as the base. The CSIQ-IQA dataset comprises 866 images that have been synthetically distorted using six types of distortions on top of 30 reference images. Lastly, the KADID dataset comprises 10125 images synthetically distorted using 25 different distortions on 81 reference images.

\subsection{Training Configurations}

Our content-aware encoder is pre-trained on the ImageNet database following the configuration proposed in MoCo-v2. Due to time and resource constraints, we train the content-aware encoder for 200 epochs. For the quality-aware encoder, we used ResNet-50 as a feature extractor, and a 2-layer MLP head to regress contrastive features of dimension $128$. The hidden dimension of the MLP head has 2048 neurons. In each forward pass, the OverLap Area (OLA) based cropping mechanism chooses two crops ($C_1$ and $C_2$) from each image, such that the percentage of overlap between the crops is maintained within a minimum and a maximum bound. The performance variation of Re-IQA-QA (quality aware module only) and Re-IQA against percentage OverLap Area, patch-sizes, and the number of image distortion augmentations is depicted in Table \ref{tab:ablationcombined}. The optimal parameters are chosen based on the combined performance of the two modules in Re-IQA. We achieved the best performance using $10-30\%$ percentage OverLap Area bound, patch size of $160$, and 11 distortion augmentations. \\
\indent The processed chunks are passed through the query and key encoders respectively in the MoCo-v2 framework, followed by an adaptive pooling layer to compress the output of the ResNet-50 into a 1D feature vector. The generated feature vector is then fed to the MLP head to generate the contrastive feature vectors required for loss computation. Our design of the Re-IQA model is inspired by previous works \cite{Wang2003MultiscaleSS, DBLP:journals/corr/abs-2110-13266} that use images both at their original and half-scale. Therefore, during the training phase of the Re-IQA model we use all images in a database both at original and half-scale, thereby doubling the training dataset.  \\
\indent During training, the following hyper-parameters were fixed throughout all experiments: learning rate $=0.6$ with cosine annealing \cite{DBLP:journals/corr/LoshchilovH16a} scheduler, InfoNCE temperature $\tau=0.2$, and momentum coefficient $=0.999$.  Our best-performing model required a batch size of $630$ (effectively $630$ x $(n+1)$ augmentations x $(2)$ scales) during training and was trained for 25 epochs. Convergence occurs in a relatively shorter number of epochs as the effective dataset size increases drastically due to a large number of augmentations and processing of each image in the dataset at two scales. All the implementations were done in Python using the PyTorch deep learning framework. We trained the content and quality-aware encoders on a system configured with 18 Nvidia A100-40GB GPUs.

\subsection{Evaluation Protocol}
We tested our Re-IQA model against other state-of-the-art models on all of the ``In the Wild" and synthetically distorted IQA databases described in Section \ref{ssec:eval}. Each of these datasets is a collection of images labeled by subjective opinions of picture quality in the form of the mean of the opinion scores (MOS). The single-layer regressor head in Re-IQA is trained by feeding the output of the pre-trained encoders and then comparing the output of the regressor, against the ground truth MOS using L2 loss. We use both Spearman’s rank order correlation coefficient (SRCC) and Pearson’s linear correlation coefficient (PLCC) as metrics to evaluate the trained model across IQA databases. \\
\indent Following the evaluation protocol used in \cite{DBLP:journals/corr/abs-2110-13266}, each dataset was randomly divided into 70\%, 10\% and 20\% corresponding to training, validation and test sets, respectively. We used the validation set to determine the regularization coefficient of the regressor head using a 1D grid search over values in the range [$10^{-3},10^3$]. To avoid overlap of contents in datasets with synthetic distortions, splits were selected based on source images. We also prevented any bias towards the training set selection by repeating the train/test split operation 10 times and reporting the median performance. On FLIVE, due to the large dataset size, we follow the train-test split recommended by the authors in \cite{DBLP:journals/corr/abs-1912-10088}.

\begin{figure}[htbp]
    \centering
    \subfloat[Re-IQA Quality-Aware]{{\includegraphics[width=0.65\linewidth]{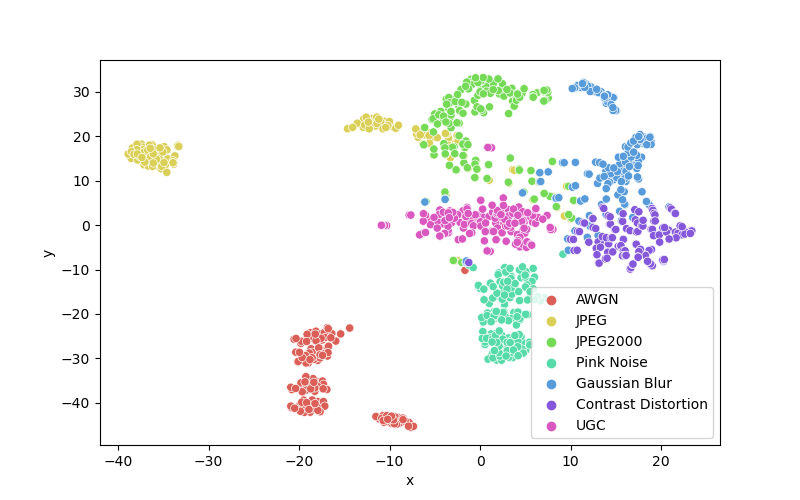}}}
    \\
    \subfloat[CONTRIQUE]{{\includegraphics[width=0.65\linewidth]{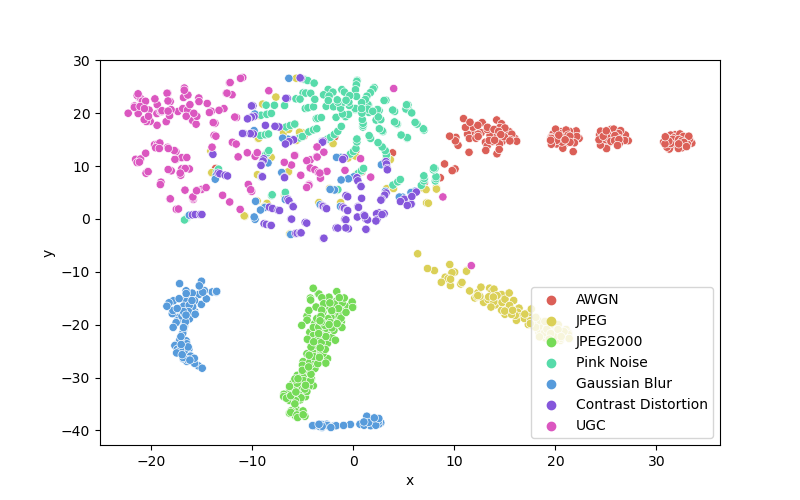}}}

    \caption{Comparison of 2D TSNE Visualization of learned representations of 1016 images sampled from KonIQ (UGC - \#150) and CSIQ (Synthetic Distortions - \#866) between Re-IQA Quality Aware sub-module and CONTRIQUE. Best viewed zoomed in.}%
    \label{fig:tsne}
\end{figure}

\subsection{Results}
Our ``Mixture of Experts" approach in Re-IQA enables us to learn robust high-level image content and low-level quality aware representations independently, the benefit of which can be clearly observed in the performance values reported in Table \ref{tab:resultsmain}. We compared the performance of Re-IQA along with its sub-modules against other state-of-the-art models on IQA datasets containing authentic and synthetic distortions in Table \ref{tab:resultsmain}. \textcolor{black}{We used the features extracted from the Resnet-50 backbone for the supervised Imagenet pre-trained model.} From the results, we conclude that Re-IQA achieves competitive performance across all tested databases. \\
\indent Results from Table \ref{tab:resultsmain} highlight the impact of content and low-level image quality on the final NR-IQA task. We observe that high-level content-aware features dominate quality-aware features for authentically distorted images, while the quality-aware features dominate the high-level content-aware features for synthetically distorted images. We can hypothesize the reason to be high variation in content in the ``Images in the Wild" scenario. Training a simple linear regressor head that is fed with features from both the content and quality-aware encoders provides flexibility to adjust the final model based on the application dataset. This can be observed in the performance scores achieved by the combined model when compared to the individual sub-modules. The performance scores of the quality-aware sub-module do not beat other methods when considering the ``Images in the Wild'' scenario, primarily due to the heavy impact of content. Despite this, when evaluated on synthetic distortion datasets, the quality-aware sub-module of Re-IQA outperforms most of its competitors all by itself. Thus we conclude that our generated quality-aware representations align very well with distortions present in an image. This is also conclusive from the t-SNE visualizations \cite{van2008visualizing} depicted in Figure \ref{fig:tsne}. \textcolor{black}{Further details on t-SNE experiments are shared in Section \ref{ssec:visl}}.

\section{Concluding Remarks}

We developed a holistic approach to Image quality Assessment by individually targeting the impact of content and distortion on the overall image quality score. NR-IQA for ``Images in the Wild" benefits significantly from content-aware image representations, especially when learned in an unsupervised setting. This work aims to demonstrate that complementary content and image quality-aware features can be learned and, when combined, achieve competitive performance across all evaluated IQA databases. We re-engineer the MoCo-v2 framework for learning quality-aware representations to include our proposed novel Image Augmentation, OLA-based smart cropping, and a Half-Swap scheme. The results of experiments on the eight IQA datasets demonstrate that Re-IQA can consistently achieve state-of-the-art performance. Our Re-IQA framework is flexible to changes in the design of encoder architectures and can be extended to other CNN architectures and Transformer based models like MUSIQ. Although developed for IQA tasks, Re-IQA can be extended as a spatial feature extraction module in Video Quality Assessment algorithms that currently use supervised pre-trained Resnet-50. 

\section{Acknowledgement}
The authors acknowledge the Texas Advanced Computing Center (TACC), at the University of Texas at Austin for providing HPC, visualization, database, and grid resources that have contributed to the research results reported in this paper. URL: \href{http://www.tacc.utexas.edu}{http://www.tacc.utexas.edu}

 \section{Appendix}
 \subsection{Distortion Bank} \label{ssec:dbanks}

\begin{figure*}[t!]
    \centering
    {\includegraphics[width=1.0\textwidth]{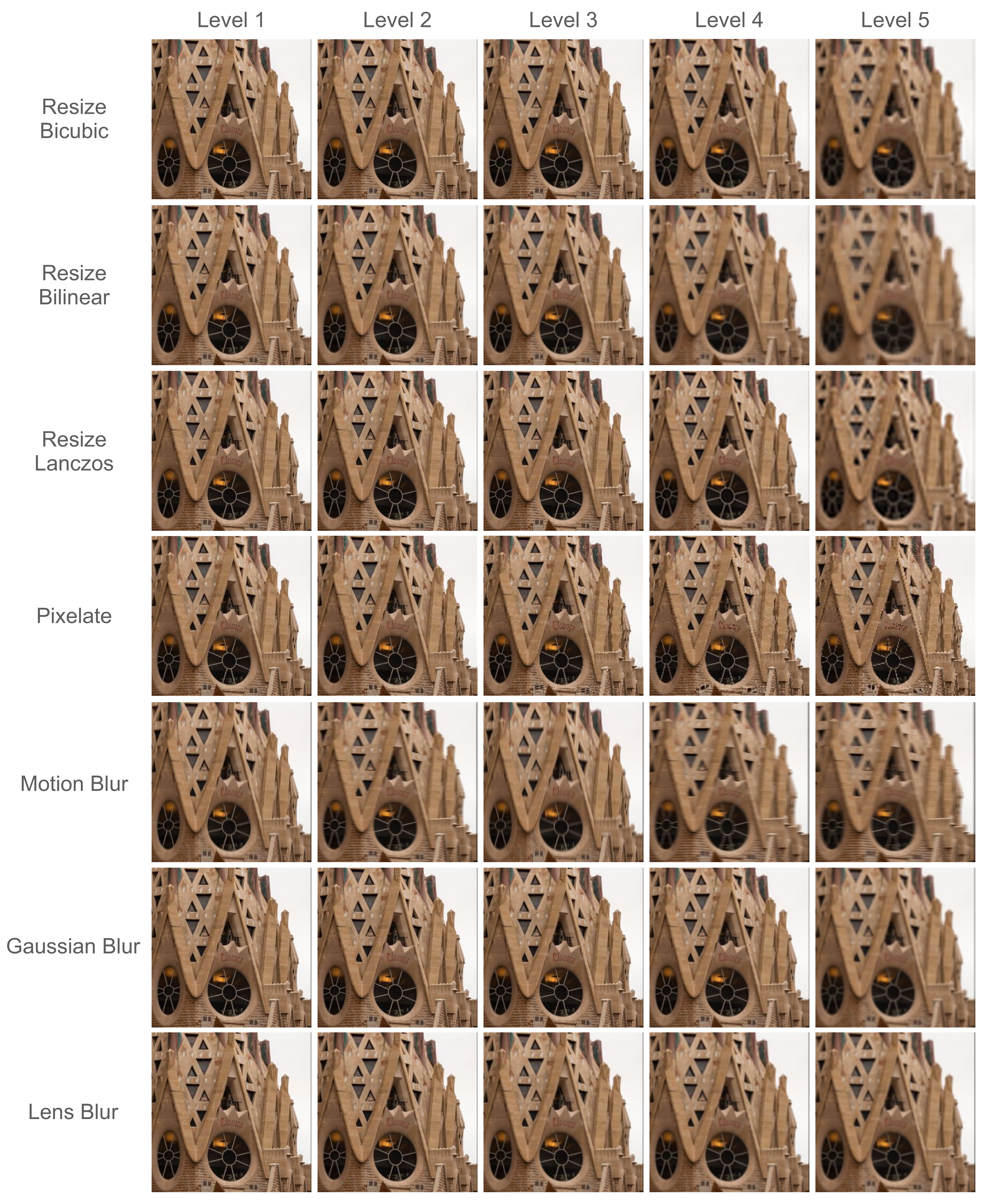}}
    \captionsetup{justification=justified}
    \caption{Distortion Bank. Best viewed zoomed.}
    \label{fig:distpage1}
\end{figure*}

\begin{figure*}[t!]
    \centering
    {\includegraphics[width=1.0\textwidth]{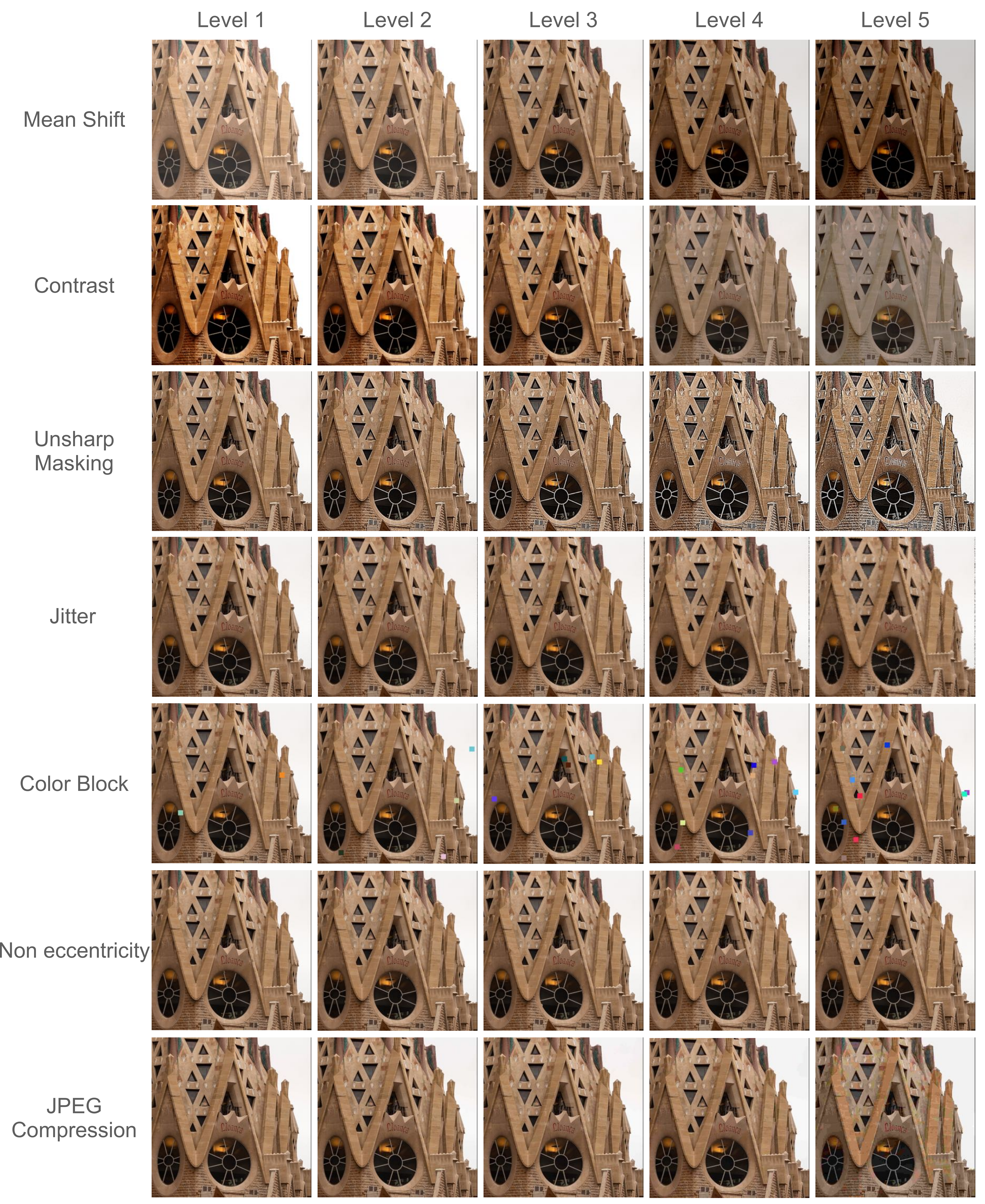}}
    \captionsetup{justification=justified}
    \caption{Distortion Bank - Continued. Best viewed zoomed.}
    \label{fig:distpage2}
\end{figure*}

\begin{figure*}[t!]
    \centering
    {\includegraphics[width=1.0\textwidth]{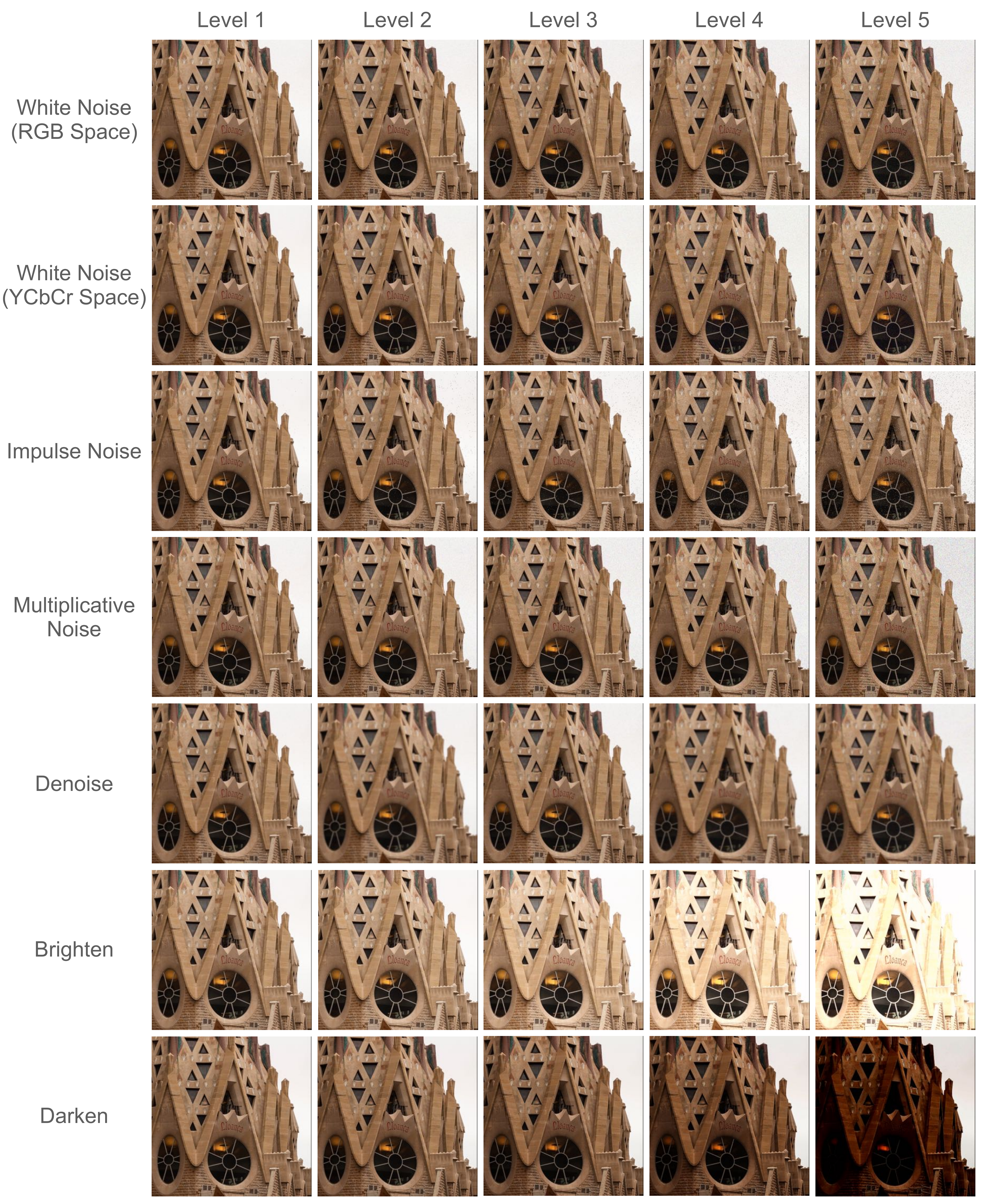}}
    \captionsetup{justification=justified}
    \caption{Distortion Bank - Continued. Best viewed zoomed.}
    \label{fig:distpage3}
\end{figure*}

\begin{figure*}[t!]
    \centering
    {\includegraphics[width=1.0\textwidth]{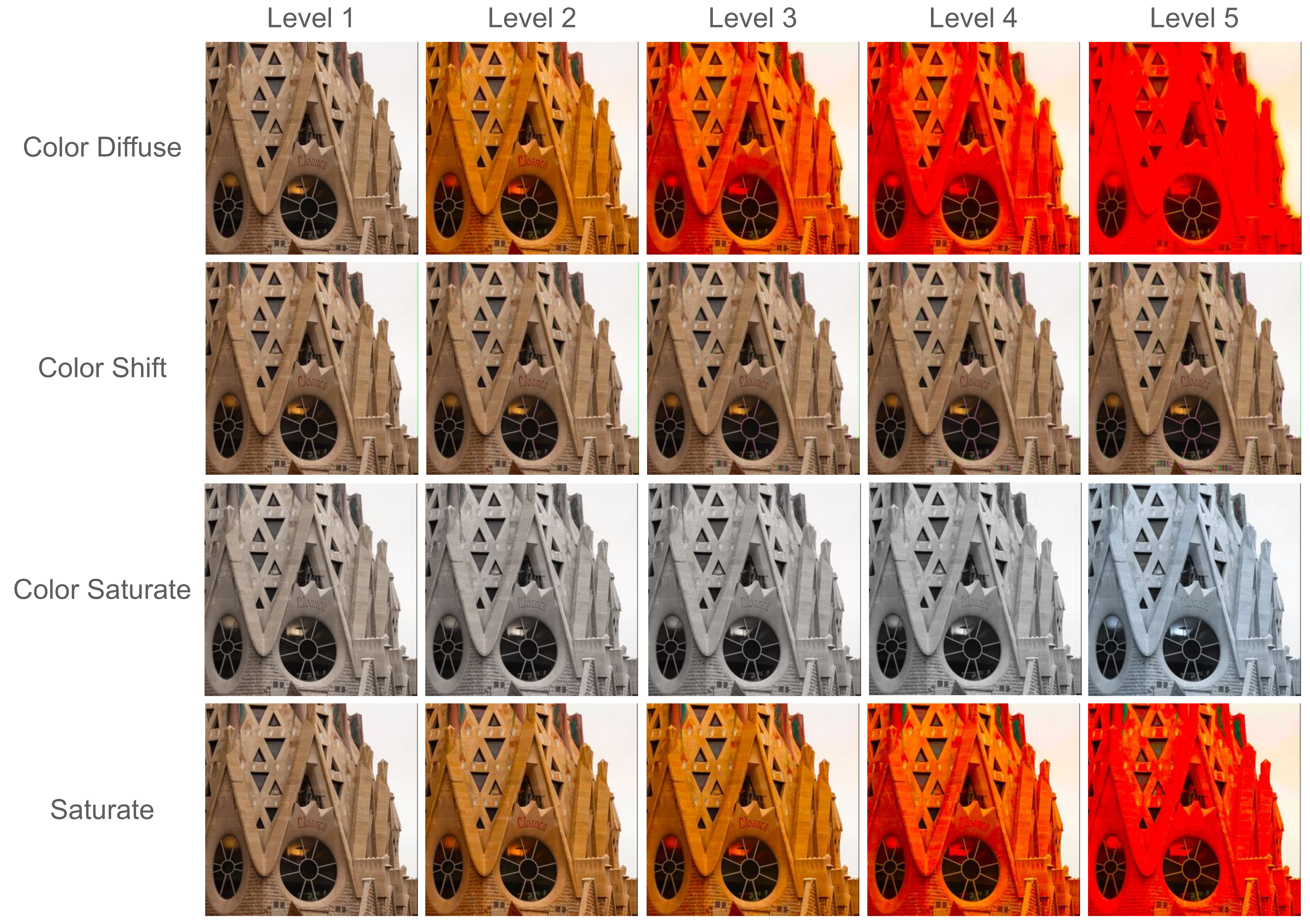}}
    \captionsetup{justification=justified}
    \caption{Distortion Bank - Continued. Best viewed zoomed.}
    \label{fig:distpage4}
\end{figure*}

We used a total of 25 quality distortion augmentations, with five levels for each type of distortion, for training our quality-aware sub-module. The visual differences among the various levels and types of distortions can be seen in Figure \ref{fig:distpage1}, \ref{fig:distpage2}, \ref{fig:distpage3}, and \ref{fig:distpage4}. The details of each distortion type are expanded below.

\begin{itemize}
  
\item \texttt{Resize Bicubic}: Downsize the image and upsize it back to the original size using bicubic interpolation.
\item \texttt{Resize Bilinear}: Downsize the image and upsize it back to the original size using bilinear interpolation.
\item \texttt{Resize Lanczos}: Downsize the image and upsize it back to the original size using Lanczos filter-based interpolation.
\item \texttt{Pixelate}: Downsize the image and upsize it back to the original size using nearest-neighbor interpolation.
\item \texttt{Motion Blur}: Emulates motion blur by filtering using a line kernel.
\item \texttt{Gaussian Blur}: Filters the image with a Gaussian kernel.
\item \texttt{Lens Blur}: Filters the image with a circular kernel.
\item \texttt{Mean Shift}: Shifts the mean intensity of the image by adding a constant value to all pixel values and truncating to the original value range.
\item \texttt{Contrast}: Changes the contrast of the image by applying a non-linear Sigmoid-type adjustment curve on the RGB values.
\item \texttt{Unsharp Masking}: Increases the sharpness of an image by using unsharp masking.
\item \texttt{Jitter}: Randomly scatters image data by warping each pixel with small random offsets.
\item \texttt{Color Block}: Inserts homogenous randomly colored blocks at random locations in the image.
\item \texttt{Non-eccentricity}: Randomly offsets small patches in the image by small displacements.
\item \texttt{JPEG Compression}: Applies standard JPEG compression.
\item \texttt{White Noise (RGB space)}: Adds Gaussian White noise in the RGB space.
\item \texttt{White Noise (YCbCr space)}: Adds Gaussian white noise in the YCbCr space.
\item \texttt{Impulse Noise}: Adds salt and pepper noise in the RGB space.
\item \texttt{Multiplicative Noise}: Adds speckle noise in the RGB space.
\item \texttt{Denoise}: Adds Gaussian white noise to the RGB image and then applies a randomly chosen blur filter (Gaussian or box blur) to remove noise.
\item \texttt{Brighten}: Increases the brightness of the image by applying a non-linear curve fitting to avoid changing extreme values.
\item \texttt{Darken}: Similar to \texttt{Brighten}, but decreases pixel values.
\item \texttt{Color Diffuse}: Applies Gaussian blur on the color channels (a and b) in the LAB-color space.
\item \texttt{Color Shift}: Randomly translates the green channel and blends it into the original image masked by a gray level map which is the normalized gradient magnitude of the original image.
\item \texttt{Color Saturate}: Multiplies the saturation channel in the HSV-color space by a factor.
\item \texttt{Saturate}: Multiplies the color channels in the LAB-color space by a factor.

\end{itemize}

\subsection{Additional Experiments}
In this section, we report results from additional experiments performed using our proposed Re-IQA framework. Section \ref{ssec:cross} illustrates the robust performance of our Re-IQA model when trained and tested on different databases. In Section \ref{ssec:fr}, we extend the Re-IQA framework to the Full Reference IQA setting and report performance results for databases with synthetic distortions.


\subsubsection{Cross-Database Performance}
We conducted cross-dataset evaluations to demonstrate the robustness of the representation learned in our Re-IQA framework, where training and testing were performed on different datasets. We follow the cross-database evaluation protocol, used by authors of CONTRIQUE. We select four image quality databases CLIVE, KonIQ, LIVE-IQA and CSIQ-IQA, covering both synthetic and authentic distortions and three state-of-the-art NR-IQA algorithms PQR, HyperIQA, and CONTRIQUE, apart from our proposed Re-IQA. We use Spearman’s rank order correlation coefficient (SRCC) to evaluate the cross-database performance of all the four models. Results from Table \ref{tab:crossdbmy-table} indicate that Re-IQA attains comparable performance to the NR-IQA models across both synthetic and authentic distortions.
\label{ssec:cross}

\begin{table}[]
\centering
\resizebox{\columnwidth}{!}{%
\begin{tabular}{c|c|cccc}
\hline
\multirow{2}{*}{Training Database} & \multirow{2}{*}{Testing Database} &  & \multicolumn{3}{c}{NR-IQA Algorithms} \\
         &          & PQR   & Hyper-IQA & CONTRIQUE & Re-IQA \\ \hline
CLIVE    & KonIQ    & 0.757 & \textbf{0.772}     & 0.676     & \textbf{0.769}  \\
KonIQ    & CLIVE    & 0.770 & \textbf{0.785}     & 0.731     & \textbf{0.791}  \\
LIVE-IQA & CSIQ-IQA & 0.719 & 0.744     & \textbf{0.823}     & \textbf{0.808}  \\
CSIQ-IQA & LIVE-IQA & 0.922 & \textbf{0.926}     & 0.925    & \textbf{0.929}  \\ \hline
\end{tabular}%
}
\caption{Cross Database SRCC Comparison of  NR-IQA models. Top 2 Performing Models in each row is in bold. Higher SRCC indicates better performance. }
\label{tab:crossdbmy-table}
\end{table}

\begin{table*}[t]
\small	
\centering
\begin{tabular}{c|cccccccc}
\hline
\multirow{3}{*}{Method} & \multicolumn{8}{c}{Synthetic Distortions}                                          \\ \cline{2-9} 
 & \multicolumn{2}{c}{LIVE-IQA} & \multicolumn{2}{c|}{CSIQ-IQA} & \multicolumn{2}{c}{TID-2013} & \multicolumn{2}{c}{KADID} \\ \cline{2-9} 
                        & SRCC  & PLCC  & SRCC  & \multicolumn{1}{c|}{PLCC}  & SRCC  & PLCC  & SRCC  & PLCC  \\ \hline
PSNR                    & 0.881 & 0.868 & 0.820 & \multicolumn{1}{c|}{0.824} & 0.643 & 0.675 & 0.677 & 0.680 \\
SSIM                    & 0.921 & 0.911 & 0.854 & \multicolumn{1}{c|}{0.835} & 0.642 & 0.698 & 0.641 & 0.633 \\
FSIM                    & 0.964 & 0.954 & 0.934 & \multicolumn{1}{c|}{0.919} & 0.852 & 0.875 & 0.854 & 0.850 \\
LPIPS                   & 0.932 & 0.936 & 0.884 & \multicolumn{1}{c|}{0.906} & 0.673 & 0.756 & 0.721 & 0.713 \\
DRF-IQA                 & \textbf{0.983} & \textbf{0.983} & \textbf{0.964} & \multicolumn{1}{c|}{0.960} & \textbf{0.944} & \textbf{0.942} & -     & -     \\
CONTRIQUE-FR            & 0.966 & 0.966 & 0.956 & \multicolumn{1}{c|}{\textbf{0.964}} & 0.909 & 0.915 & \textbf{0.946} & \textbf{0.947} \\ \hline
Re-IQA-FR (content-aware)  & 0.932 & 0.947 & 0.927 & \multicolumn{1}{c|}{0.928} & 0.877 & 0.884 & 0.889 & 0.898 \\
Re-IQA-FR (quality-aware)  & \textbf{0.973} & \textbf{0.974} & \textbf{0.961} & \multicolumn{1}{c|}{\textbf{0.965}} & 0.905 & 0.915 & 0.901 & 0.903 \\
Re-IQA-FR   & 0.969 & \textbf{0.974} & \textbf{0.961} & \multicolumn{1}{c|}{0.962} & \textbf{0.920} & \textbf{0.921} & \textbf{0.933} & \textbf{0.936}
\end{tabular}
\caption{Performance comparison of Re-IQA-FR against various FR-IQA models on IQA databases with synthetic distortions. The top 2 best performing models are in bold. Higher SRCC and PLCC scores imply better performance. }
\label{tab:fr}
\end{table*}
\subsubsection{Full Reference Image Quality Assessment}
\label{ssec:fr}

Re-IQA learns image representations in a completely  unsupervised setting, thus making it flexible for application in Full Reference Image Quality Assessment (FR-IQA) tasks. Since reference image scores are available in the case of FR-IQA tasks, the score regressor is now trained to predict the difference in the scores of the reference image and the distorted image, as described in equation \ref{eq:FR-IQA} below

\begin{equation}
    y_{dmos} = W|h_{ref}-h_{dist}|
    \label{eq:FR-IQA}
\end{equation}

where $y_{dmos}$ is the differential mean opinion score of the test image, $W$ represents the regressor weights, and $h_{ref}$ and $h_{dist}$ are the image representations generated by the Re-IQA sub-modules for the reference and distorted images respectively. We refer to this modified framework as Re-IQA-FR. Note that only the score regressor is modified and trained for FR-IQA, and the weights of the pre-trained sub-modules of Re-IQA remain unchanged. \\
\indent The evaluation strategy for FR-IQA remains exactly the same as that of NR-IQA, wherein we split the dataset into 70-10-20 for train-val-test sets and report the median performance scores. We report the performance of Re-IQA-FR on various full reference IQA datasets (refer Table \ref{tab:fr}), otherwise referred to as synthetic IQA datasets throughout our paper. We compare our model against six popular FR-IQA models available in the literature: PSNR, SSIM, FSIM, LPIPS, DRF-IQA \cite{10.1109/TIP.2020.2968283} and CONTRIQUE-FR, where the first three are traditional models and the last three are deep learning based models. We also report the scores of the individual sub-modules of Re-IQA-FR. \\
\indent Results from Table \ref{tab:fr} highlight that our proposed Re-IQA-FR model achieves state-of-art performance across all evaluated databases. Another interesting observation is that the SRCC and PLCC scores of the Re-IQA-FR are better than the No-Reference version of Re-IQA across all databases, as reported in Table \ref{tab:resultsmain}. This can be attributed to the knowledge of extra information available to the Re-IQA-FR framework in the form of the undistorted reference image. We also provide a cross-database comparison between CONTRIQUE-FR and Re-IQA-FR in Table \ref{tab:crossfr-table}. The results suggest that RE-IQA-FR and CONTRIQUE-FR exhibit similar performance, with RE-IQA-FR performing slightly better.

\begin{table}[t]
\centering
\caption{Cross Database SRCC Comparison of CONTRIQUE and Re-IQA-FR}
\resizebox{\columnwidth}{!}{%
\begin{tabular}{c|c|cc}
\hline
Training Dataset & Testing Dataset  & CONTRIQUE-FR   & Re-IQA-FR      \\ \hline
LIVE-IQA         & CSIQ-IQA       & 0.855          & 0.856 \\
CSIQ-IQA         & LIVE-IQA           & 0.937 & 0.942    
\end{tabular}%
}
\label{tab:crossfr-table}
\end{table}

\subsection{Visualization} \label{ssec:visl}

\begin{figure}[t!]
    \centering
    {\includegraphics[width=1.0\linewidth]{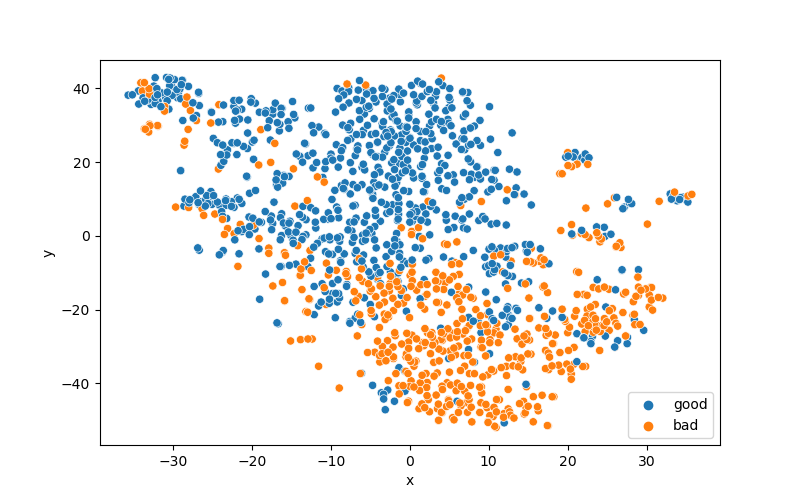}}
    \captionsetup{justification=justified}
    \caption{t-SNE visualization of learned quality-aware representations for 1338 `Images in the Wild' from KonIQ.}
    \label{fig:TSNE-ugc}
\end{figure}

We utilized the TSNE() function from the sklearn.manifold library to conduct t-SNE experiments and present the findings in Figure \ref{fig:tsne}. The function's default parameters were used, except for $perplexity$ and $n\_iter$. While we selected $perplexity=50$ and $n\_iter=3000$ to achieve optimal visual outcomes, our results remain reliable even when $perplexity=10$ and $n\_iter=1000$ were used.\\
In Figure \ref{fig:TSNE-ugc}, we demonstrate the efficacy of the representations generated by the quality-aware sub-module of Re-IQA in segregating high-quality images from low-quality images. We use 2D t-SNE  plots to visualize the high-dimensional learned feature embeddings. The orange points represent lower-quality images (scored less than 50 on a scale of 0-100), while the blue points represent higher-quality images (scored more than 50 on a scale of 0-100). We chose 1338 images from the KonIQ dataset, such that there is an equal representation of low and high-quality images available for the visualization. As can be seen in the figure, the orange points cluster toward one corner, whereas the blue points cluster toward the opposite corner. The overlapping region of the clusters represents those images for which the quality-aware model's learned representations are likely to produce quality scores that do not agree with the ground truth.

\subsection{Discussion on Hypotheses} \label{ssec:dischyp}
Our unsupervised learning framework to learn quality-aware features requires us to define a decision boundary between `similar-quality' and `different-quality' labels. As discussed in Section \ref{ssec:QAIR},  we set up our learning framework by assuming the following three hypotheses to be true:\\ \\
 \textbf{Hyothesis 1}: Image quality can vary in different locations based on the image's content. Thus, PQAF varies within an image itself. If we assign PQAF to an image patch $x$ and denote it as $PQAF_{x}$, then $PQAF_{x}$ varies only a small amount between neighboring patches. However, $PQAF_{x}$ may vary significantly between two distant patches depending on the content. Our approach generates `similar-quality' samples by extracting overlapping patches from an image in the training database. We conducted an ablation study to assess the impact of varying the degree of overlap between the two patches sampled from an image on the performance of the Re-IQA model. Detailed results of this study can be found in Table \ref{tab:ablationcombined}.  We do not explicitly sample from distant patches in an image.  \\ \\
 \textbf{Hypothesis 2}: This hypothesis serves as a primary basis in the design of the Re-IQA-QA module. The PQAF of any two randomly selected images are `different,' which assumes that the scenes depicted in the images differ. However, this does not enforce any restrictions on the quality scores of the two images. Without this hypothesis, the contrastive loss cannot be computed between features obtained from crops of 2 different source images, and the network will not learn to compare two images from different sources during training. \\ \\
\textbf{Hypothesis 3}:  Two different distorted versions of the same image have different PQAF. This hypothesis is like an axiom. Its absence results in $2$ images with the same content but different distortions generating the same PQAF, which would be technically wrong because it would imply two images with different distortions have the same perceptual quality. This is important because it ensures that the PQAF accurately reflects the perceptual quality of an image, even when comparing images with the same content but different distortions. \\

\section{Change Log}
\begin{itemize}
    \item v1: April 02, 2023, First version uploaded to Arxiv.
    \item v2: May 27, 2023, Minor Typo fixed.
\end{itemize}




{\small
\bibliographystyle{ieee_fullname}
\bibliography{arxiv}
}

\end{document}